\documentclass{article}
\usepackage[11pt]{moresize}
\usepackage{graphicx}
\usepackage{subfigure}
\usepackage[british]{babel}
\usepackage{csquotes}
\usepackage[
    backend=biber,
    style=apa,
    natbib,
    sorting=ynt,
    sortcase=false,
    sortcites
  ]{biblatex}
\usepackage{authblk}
\usepackage{amsmath}
\usepackage{amssymb}
\usepackage[export]{adjustbox}
\usepackage{booktabs}
\usepackage[usenames,dvipsnames]{color}
\usepackage{siunitx}  
\DeclareSIUnit\pixel{px}

\usepackage[colorlinks=true,allcolors=black]{hyperref}

\usepackage[switch]{lineno}
\usepackage{hyphenat}
\begin{document}

\title{Precise localization of corneal reflections in eye images using deep learning trained on synthetic data}

\author[1]{
Sean Anthony Byrne}
\author[2]{Marcus Nystr{\"o}m}
\author[3]{Virmarie Maquiling}
\author[3]{Enkelejda Kasneci}
\author[2,4]{Diederick C. Niehorster}

\affil[1]{MoMiLab, IMT School for Advanced Studies Lucca, Lucca, Italy}
\affil[2]{Lund University Humanities Lab, Lund University, Lund, Sweden}
\affil[3]{Human-Centered Technologies for Learning, Technical University of Munich, Munich, Germany}
\affil[4]{Department of Psychology, Lund University, Lund, Sweden}

\affil[ ]{\texttt{Sean.byrne@imtlucca.it}, \texttt{marcus.nystrom@humlab.lu.se }, \texttt{virmarie.maquiling@tum.de},
\texttt{enkelejda.kasneci@tum.de},
\texttt{diederick$\_$c.niehorster@humlab.lu.se*}}


\maketitle
\begin{abstract}
We present a deep learning method for accurately localizing the center of a single corneal reflection (CR) in an eye image. Unlike previous approaches, we use a convolutional neural network (CNN) that was trained solely using synthetic data. Using only synthetic data has the benefit of completely sidestepping the time-consuming process of manual annotation that is required for supervised training on real eye images.
To systematically evaluate the accuracy of our method, we first tested it on images with synthetic CRs placed on different backgrounds and embedded in varying levels of noise. Second, we tested the method on two datasets consisting of high-quality videos captured from real eyes.
Our method outperformed state-of-the-art algorithmic methods on real eye images with a \SIrange{13}{41.5}{\percent}reduction in terms of spatial precision across data sets, and performed on par with state-of-the-art on synthetic images in terms of spatial accuracy. We conclude that our method provides a precise method for CR center localization and provides a solution to the data availability problem which is one of the important common roadblocks in the development of deep learning models for gaze estimation. Due to the superior CR center localization and ease of application, our method has the potential to improve the accuracy and precision of CR-based eye trackers.
\end{abstract}

\section{Introduction}

An important part of the image processing pipeline in many video-based eye trackers is to localize the center of certain features in the eye image, typically the pupil \cite{fuhl2017pupilnet,1565386} and one or multiple corneal reflections (CRs)\cite{perez2003precise,Nystroem2022}. Accurate localization of these features is a prerequisite for an accurate gaze signal produced by the eye tracker. In this paper we focus on localizing the center of a single CR, which together with the pupil (P) center constitute the input for the dominant principle for video-based eye tracking over the past decades \cite{merchant1974remote}. This principle is known as P-CR eye tracking, and is used in high-end commercial systems like the EyeLink from SR Research (Ontario, Canada), where the head is typically constrained relative to the eye camera with a chin and forehead rest.

Conventionally, researchers have relied on using algorithmic methods which require a series of hard coded steps to yield a feature center estimate from the input image. Recent work has however shown that these traditional methods are limited in how accurately a pupil or CR can be localized in an eye image, in particular in the presence of image noise \cite{Nystroem2022}.
To overcome these challenges, we introduce a deep learning method that---trained on a set of synthetic eye images---is able to accurately locate CR centers in real-world eye images. From a deep learning perspective, using synthetic data tackles the longstanding problem of finding or creating large, annotated databases that deep learning models are traditionally trained on. Our approach not only outperforms the traditional algorithmic methods used by many commercial eye-trackers,
but also introduces a new paradigm for 
researchers interested in training deep learning models for gaze estimation. Our approach is not only simple to train, but also offers the flexibility to change the synthetic images used for training to suit the specific needs of the user. Additionally, our approach eliminates the need for large datasets and the time-consuming act of hand labeling images, which are often impractical for smaller research labs due to time and financial constraints.

There are several algorithmic methods to detecting CRs in eye images and localizing their center, such as thresholding followed by center of mass calculations or ellipse fitting \cite{Nystroem2022,shortis1994comparison}. Many commercial eye trackers use these approaches, however they tend to suffer from inaccuracies and poor resolution, such as mis-estimating the amplitude of small eye movements \cite{holmqvist2020small}. Based on their results, \citet{holmqvist2020small} hypothesized that these problems could be ascribed to mislocalization of the CR center. Recent theoretical work has indeed shown that accurate localization of the center of the CR with thresholding techniques requires that the CR spans a sufficient number of pixels in the eye images and that the level of image noise remains low~\cite{Nystroem2022}. Besides simple thresholding of the CR in the eye image, other algorithmic methods for CR center localization have been proposed. Recently, \citet{Wu2022} showed that radial symmetry \cite{parthasarathy2012rapid}, a particle tracking method stemming from the field of microscopy, outperformed simple thresholding in terms of CR localization. However, they only tested the radial symmetry method on eye images where the CR was located fully inside the pupil and thus appeared on a uniform dark background. Depending on the range of gaze directions in the task and the geometry of the setup, the CR will also often be located on the iris or on the border between the pupil and the iris. Since this case was not evaluated by \citet{Wu2022} who only used a limited range of gaze directions, it remains unclear how the radial symmetry method would perform on real eye images recorded during unconstrained viewing.


In this study, we aimed to improve the accuracy and precision of high-end P-CR eye tracking in controlled laboratory settings to allow even more robust estimation of small and slow eye movements. We use a two-stage approach, where a traditional method is used for rough localization, followed by a CNN-based method for more accurate localization. Specifically, to simplify the training procedure, and since traditional algorithmic methods such as thresholding are already quite good at CR localization, we first use thresholding followed by centroid calculation to make an initial estimate of CR center location, and then use the CNN on image patches centered around these locations. We compared the performance of our deep learning method to traditional methods such as radial symmetry and thresholding. We predicted that a CNN, when trained effectively, will be able to surpass traditional methods and achieve improved localization performance 
\cite[\emph{cf.},][]{helgadottir2019digital}.
In order to train this model effectively we synthesized CRs using 2D Gaussian distributions with different sizes and positions on a background with varying levels of noise (see the ``\nameref{sec:methods_sim}'' section below for a detailed description). 

We use a deep learning framework known as DeepTrack \cite{helgadottir2019digital}, originally developed for particle tracking in a microscopy setting. This framework utilizes a CNN trained on synthetic data to track single particles and also includes a U-Net model for tracking multiple particles. Subsequent work using this framework has also incorporated single-shot self-supervised object detection and geometric deep learning models \cite{midtvedt2022single,https://doi.org/10.48550/arxiv.2202.06355}. The DeepTrack 2.1 Python library \cite{midtvedt2021quantitative} makes it easy to generate a synthetic dataset and train a deep learning model in the same pipeline. To the best of our knowledge, this is the first time the DeepTrack 2.1 library has been used outside of digital microscopy. 

While many gaze estimation studies deploy deep learning architectures, we focus our literature review specifically on models whose aim is to locate the CR or who create synthetic data used to train the model. We could find only three other deep learning models that have been developed for locating CRs in eye images. First, \citet{wu2019eyenet} used a three-level CNN network to locate CRs and match their locations in the eye image to the physical locations of the light sources used the generate the CRs. The model included a CNN backbone with feature pyramid outputs as the base architecture. The output was passed into two additional networks, one for identity matching 
and another for localization. Second, \citet{chugh2021detection} trained a U-Net model on 4000  hand-labeled real eye images to accurately detect multiple CRs within the same image. The authors relied on several data augmentation techniques to develop a total of 40,000 samples for training. Third, \citet{niu2021real} proposed a lightweight model that both localizes and matches corneal reflections. This model employs an attention mechanism to identify both the pupil center and CRs. They demonstrated improved performance in CR localization and matching when compared to \citet{chugh2021detection}. It should be noted that by design, the work of \citet{wu2019eyenet} and \citet{niu2021real} was limited to localizing CR centers with pixel resolution, making their approaches unsuitable for our setting where we aim to recover the CR position at higher accuracy. The approach of \citet{chugh2021detection} is theoretically able to provide sub-pixel level CR localizations, but in practice achieved an average error of 1.5 pixels, and requires training the model for a specific CR size.

To overcome the need for large data sets to enable effective training of machine learning models for gaze estimation, previous studies have successfully used synthetic data~
\cite{DBLP:journals/corr/abs-2104-12668}. For instance, \citet{6909631} proposed a ``learning-by-synthesis'' approach that trained random regression forests on a dataset of 3D reconstructions of eye regions created using a patch-based multi-view stereo (MVS) algorithm. \citet{DBLP:journals/corr/WoodBZSRB15} also used a similar approach of generating photo-realistic eye images to be used for a wide range of head poses, gaze directions and illuminations to develop a robust gaze estimation algorithm, and in a separate study generated a dataset of 1 million synthetic eye images \cite{10.1145/2857491.2857492}. \citet{DBLP:journals/corr/ShrivastavaPTSW16} proposed an unsupervised learning paradigm using generative adversarial networks improving realism of a simulator's output by using unlabeled real data. They argue synthetic data may not achieve the desired performance due to a gap between synthetic and real eye image distributions. The method was tested for gaze and hand pose estimation, showing a significant improvement over using synthetic data alone.
Our study diverges from these previous approaches as we demonstrate that models can be trained to achieve high accuracy using simple, highly controllable synthetic images that are substantially different from real eye images, avoiding the need for sophisticated and time-consuming techniques such as data augmentation or reconstruction methods.

This paper addresses the following questions regarding deep learning methods for CR center localization: 1) Can a CNN trained on synthetic images perform CR center localization in real eye images? and if so, 2) Is our method able to locate the CR center more accurately than commonly used algorithmic approaches when applied to synthetic data? and 3) Does our method deliver a CR position signal with higher precision than the algorithmic approaches when applied to real eye images? We focus specifically on high resolution and high quality eye images, since accurate CR localization in high-end eye trackers is key to recording high-precision data where even the smallest and slowest of eye movements, e.g., microsaccades and slow pursuit, can be robustly distinguished from noise \cite[c.f.,][]{holmqvist2020small, Nystroem2022, niehorster2020apparent}. Nevertheless, to assess the potential generalizability of our method to lower resolution eye images, in a second experiment reported in this paper we also evaluate our approach on spatially downsampled eye images.

\section{Methods}

\subsection{Model Architecture}

\begin{figure*}
    \centering
    \includegraphics[width=\textwidth]{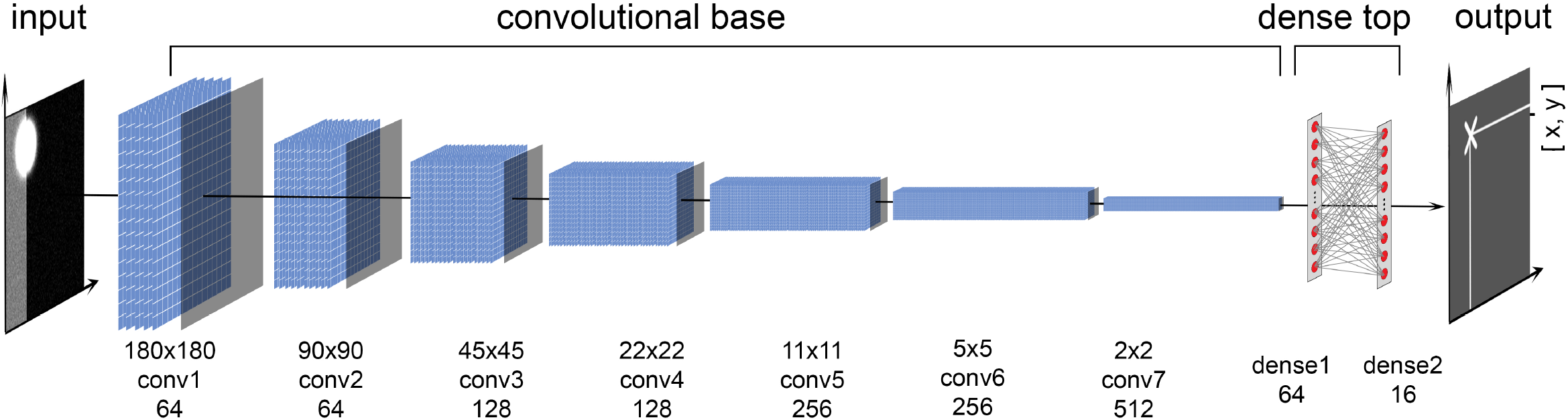}
    \caption{Overview of our method: A CNN model with seven convolutional layers that increase in filter size from 64 to 512 and two dense layers returning the Cartesian coordinates of the CR center.}
    \label{fig:model_overview}
\end{figure*}

In a preliminary test, we implemented the original DeepTrack CNN model as described in \citet{helgadottir2019digital}. This model consisted of three convolutional layers and two dense layers, and we employed the same optimizer and hyperparameter choices as described in the original work. However, when we evaluated this model on our synthetic images that simulated a corneal reflection captured in a video-based eye tracking setup, we were unable to achieve sub-pixel level accuracy. The minimum validation error we reached was 2.95 pixels. To enhance the accuracy of our predictions, we developed our own CNN model. Our model included seven convolutional layers connected to two dense layers. The input to our model is 180 x 180 pixels grayscale images, and it outputs the subpixel location of the corneal reflection center. Figure \ref{fig:model_overview} provides a visual representation of the complete architecture of our model.

\subsection{Model Training}

We implemented a two-stage training approach for our model to achieve sub-pixel level accuracy. In the first stage, to ensure good generalization, we trained the model on a broader range of CR center locations than the model would typically encounter during inference. We describe the process of generating the images in detail in the following section of the paper. We utilized the Adam optimizer \cite{kingma2017adam} and a mean squared error (MSE) loss function along with a very small batch size of four. The training was conducted for a maximum of 700 epochs, with an early stopping function implemented to prevent overfitting, achieving a validation error of 0.2338 pixels after 127 epochs. The second stage of training was performed on a dataset containing a smaller range of synthetic CR center locations. During this stage, we fine tuned the model by freezing the first two convolutional blocks while all subsequent layers of the model were set to trainable (i.e unfrozen), and we initialized the model with the weights from the first training stage. For selecting the layers to freeze, we followed an iterative process where we gradually increased the number of trainable layers. We initiated the process with a fully frozen model and subsequently unfroze the layer closest to the model head at each iteration. Additionally, we lowered the learning rate of the Adam optimiser from  $1\mathrm{e}^{-4}$ to $1\mathrm{e}^{-6}$.  The second stage of training resulted in a sub-pixel accuracy of 0.085 after 187 epochs on the validation set.

The DeepTrack 2.1 \cite{midtvedt2021quantitative} package provides a generator function which we used to efficiently generate and feed images into the model for training. We set up the generator such that the model only saw each training image one time, meaning that every image the model saw for training was unique. We additionally generated 300 synthetic images for the validation set. The fully trained model was saved and model evaluations were conducted on an Intel Xeon W-10885M CPU @ 2.40GHz with a prediction time of 13ms per image.

\subsection{Generating Synthetic Images}
\label{sec:methods_sim}

\begin{figure*}%
\includegraphics[width=\textwidth]{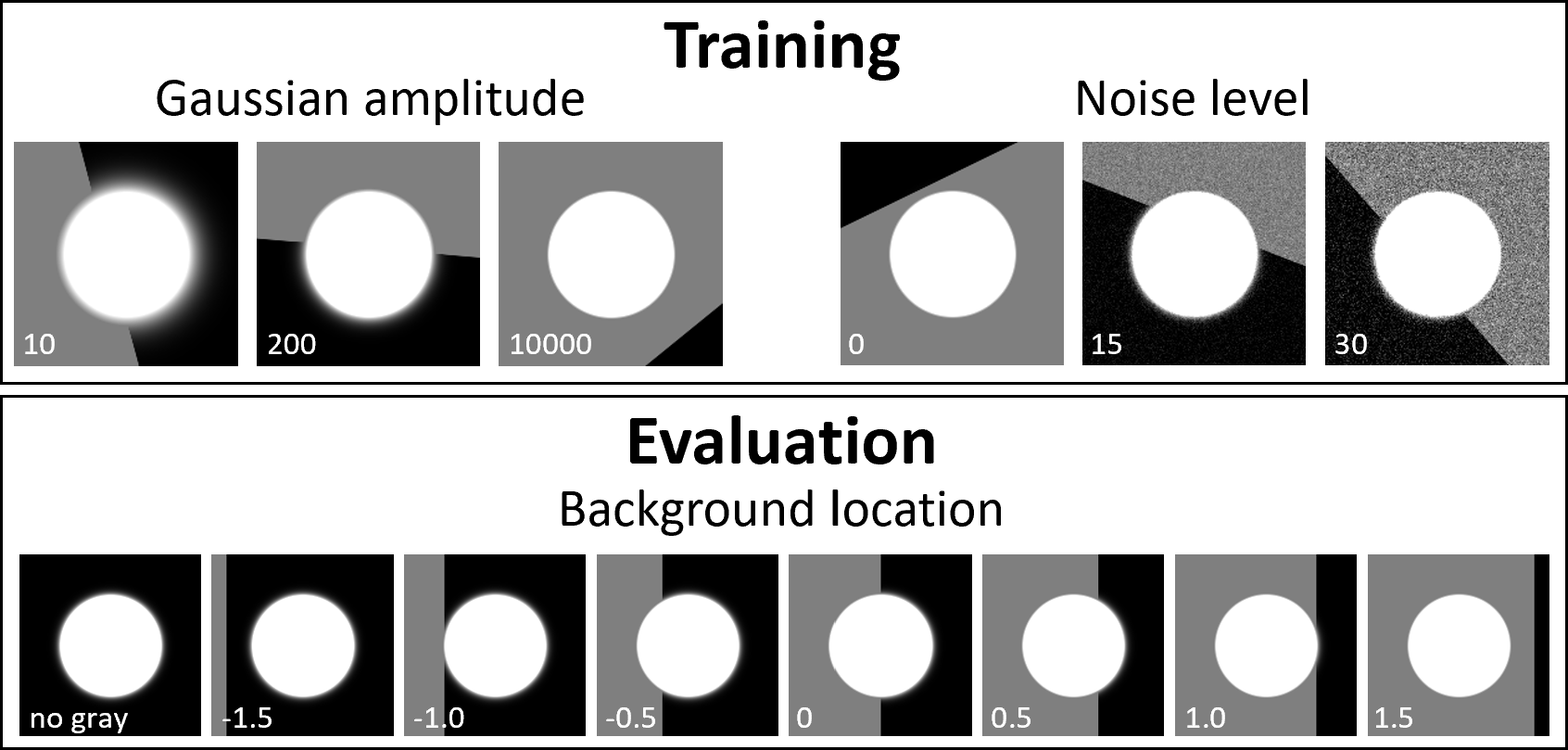}
\caption{Example simulated CRs. Top row: example images used during model training and for the validation set. Left column: different values of Gaussian amplitude $A$. Right column: different pixel noise values $\sigma_n^{2}$ (image levels). For both columns, random positions (within $[-1.5r,1.5r]$) and orientations of the dividing line between the dark and light sections of the background are shown. Bottom row: example images used for evaluation, showing different background locations $E$ as well as a CR image without a gray background. The value for the varied parameter is denoted on the panels. $A$ was set to 10000 for all panels except the top-left. For illustration purposes, the CR radius ($r$) in these panels is 50 pixels. During both training and evaluation the pixel intensity of the lighter section of the background was also varied (not shown).}%
\label{fig:CR_images}%
\end{figure*}

\subsubsection{Features of the synthetic images}
As in previous work~\cite{Nystroem2022}, the light distribution of the CR in an eye image is modeled as a 2D Gaussian distribution, as is supported by optical modeling~\cite{Wu2022}. CRs in real eye images have at least two further important features: 1) The CR in an eye image is normally heavily oversaturated~\cite{Wu2022, Holmqvist2011}; and 2) depending on the physical geometry of the setup and the orientation of the eye, the CR is often overlaid on a non-uniform background, such as the iris or the edge of the pupil. We, therefore, extend the approach of \citet{Nystroem2022} by introducing saturation that truncates the Gaussian distribution and leaves it with an area of maximum brightness surrounded by shallow tails, and by introducing a non-uniform background.

More formally, the saturated CR is generated from a Gaussian distribution
\begin{equation}
G(x,y) = Ae^{-\left(\frac{(x - x_c)^2 + (y - y_c)^2}{2 \sigma_w^2}\right)}
\label{eq:Gauss}
\end{equation}
where the following parameters are varied in the simulation:
\begin{enumerate}
    \item The center of the input light distribution $(x_c, y_c)$.
    \item The amplitude $A$ of the Gaussian distribution. Saturation is achieved when $A$ is set to amplitudes larger than 1 since image values are limited to 1 at the end of the image generation pipeline (see below). Figure \ref{fig:CR_images} (top-left) shows CRs at three different amplitudes. Notice that larger amplitudes lead to shallower tails.
    \item The radius $r$ of the resulting CR. This is parameterized such that for a given value, the radius of the saturated portion of the CR is kept constant irrespective of the amplitude ($A$) of the underlying Gaussian. This is achieved by setting $\sigma_w = r/\sqrt{-2\log \frac{1}{A}}$.
\end{enumerate}
Two further aspects were varied to generate the final synthetic images. First, to simulate the pupil-iris border, a background was generated that consisted of two sections of different luminance, and the line dividing the two sections was randomly placed near the CR and randomly oriented. The image of the synthetic CR was added to this background using the following operation: $max(CR, background)$. The top row of Figure \ref{fig:CR_images} shows synthetic CRs on various example backgrounds. Furthermore, noise was added to the images by adding a value drawn from a Gaussian distribution $X \sim \mathcal{N}(0,\,\sigma_n^{2})$ for each individual pixel of the image. The parameter $\sigma_n^{2}$ was varied (see Figure \ref{fig:CR_images}, top-right). 
Finally, the intensity values in the resulting images were limited to the range $[0, 255]$, scaled to the range $[0, 1]$ and the image was discretized to 256 levels, corresponding to 8-bit camera images.

\subsubsection{Model of image information for CR center localization}
Since the aim of the current work is to develop a high-accuracy CR center localization method, it is important to develop a model of what accuracy an optimal localization method could achieve. As shown by \cite{Nystroem2022}, CR localization accuracy depends on the number of pixels spanned by the CR in the image as well as the shape of the light distribution.
Theoretically, the lower the spatial pixel resolution or bit-depth of the CR image, the lower is the maximum achievable localization accuracy of the CR center.
This follows from the logic that the coarser the digital representation of the CR, the bigger the change in its position needs to be before an observable change occurs in the CR image \cite[\textit{cf}.,][]{mulligan1997}. Therefore, to provide a benchmark for the results presented in this paper, we determined the theoretically optimal center localization performance as a function of CR size and Gaussian amplitude (i.e. tail width). To do so, we took a set of CR images generated with the Cartesian product of $r = \{2,\allowbreak 4,\allowbreak 6,\allowbreak ...,\allowbreak 18\}$ and $A=\{10,\allowbreak 50,\allowbreak 200,\allowbreak 1000,\allowbreak 10000\}$, i.e., the same parameters as used for the evaluation on synthetic images (see the section ``\nameref{sec:sim_img_eval_methods}'' below). The center of each CR image was then estimated as the center-of-mass of all the pixels in the discretized CR image, using their intensity values~\cite{Nystroem2022,shortis1994comparison}. Unlike the synthetic images used for model evaluation, these images had a completely black background such that only pixel intensity values associated with the CR would influence the center estimate. 

\subsubsection{First \& Second Training Stages}
During the first training stage, the following parameters were used. Where possible, the parameters were set to ranges significantly larger than the set used for evaluation.
\begin{enumerate}
    \item CR radius $r$ was drawn from a uniform distribution with range $[1,30]$ pixels. This was chosen to be wider than our testing range of $[2,18]$ \cite[like was used in][]{Nystroem2022} and also encompasses the range of CR sizes one may reasonably expect to encounter in real eye images.
    \item Location: Horizontal and vertical CR center locations were drawn from uniform distributions. To ensure that the CR would not be significantly cut off by the edge of the image, the range of both uniform distributions depended on the CR size ($r$). Specifically, they spanned $[r,180-r]$ pixels, where 180 pixels is the image size.
    \item Gaussian amplitudes $A$ were drawn from a uniform distribution with range $[2,20000]$. The range of this parameter was decided by means of manual inspection of the output to provide a range of different tail widths (c.f. Figure \ref{fig:CR_images}).
    \item The horizontal and vertical coordinates of a point on the line dividing the two sections of the background were drawn from a normal distribution centered on the CR center location and spanning a standard deviation of $-1.5r$. A random orientation of this line was then drawn from a uniform distribution with range $[0,2\pi]$. The edge between the two segments was smoothed with a raised cosine profile spanning 4 pixels. The pixel intensity value of the dark section of the background was drawn from an exponential distribution with its scale parameter set to 10 pixel intensity values, and offset 1 (so that full black did not occur). The pixel intensity level of the lighter section $I$ was drawn from a uniform distribution with a range of $[32,153]$ pixel intensity values.
    \item The pixel noise $\sigma_n$ was drawn from a uniform distribution with range $[0,30]$ pixel intensity values.
\end{enumerate}

In the second training stage, all parameters except CR location were set to the same ranges as were used in the first stage. Since the CNN will only be used on image patches where the CR has already been centered, horizontal and vertical CR center locations in this second pass were drawn from uniform distributions with ranges spanning 1.5 pixels around the center of the output image, i.e. $[89.25,90.75]$.

\subsection{Evaluation}

\subsubsection{Evaluation on synthetic images}
\label{sec:sim_img_eval_methods}
To investigate how accurately the center of the CR can be located by the various methods, an input light distribution with horizontal center $x_c$ was moved in small steps ($\delta_{x_c} = 0.01$) over a one-pixel range (100 steps). The input position was then compared to the output of three methods: 1) the traditional thresholding method \cite{Nystroem2022}; 2) the radial symmetry algorithm of \citet{parthasarathy2012rapid}; and 3) the CNN developed in this paper. A fourth method which simply computes the center of mass of all the pixels in the input image \cite[called the intensity-based method in][]{Nystroem2022} was discarded after initial investigation since this method produced very large errors on our evaluation images due to the partially grey background. 

The evaluation was performed at several CR radii $r$, Gaussian amplitudes $A$, gray background locations $E$, pixel intensity values $I$ of the lighter part of the background and noise levels $\sigma_n$. Specifically, the testing set consisted of the Cartesian product of $r = \{2,\allowbreak 4,\allowbreak 6,\allowbreak ...,\allowbreak 18\}$, $A=\{10,\allowbreak 50,\allowbreak 200,\allowbreak 1000,\allowbreak 10000\}$, $\sigma_n = \{0,\allowbreak 2,\allowbreak 4,\allowbreak ...,\allowbreak 18\}$, grey background locations $E = \{no gray,\allowbreak -1.5,\allowbreak -1,\allowbreak -.5,\allowbreak 0,\allowbreak 0.5,\allowbreak 1,\allowbreak 1.5\}$, and pixel intensity levels of the lighter background section 
$I = \{38,\allowbreak 51,\allowbreak 64,\allowbreak 77,\allowbreak 89,\allowbreak 102,\allowbreak 115,\allowbreak 128,\allowbreak 140,\allowbreak 153\}$, i.e., it included all combinations of these parameters. At each combination of parameters, the horizontal center of the CR $x_c$ was moved through 100 steps of $\delta_x = 0.01$ pixels as described above. The dividing line between the two sections of the background was always vertical and it was positioned relative to the synthetic CR such that $E=0$ meant the border between the two sections coincided with the CR center, $E=-1$ that it was placed 1 CR radius $r$ to the left of the CR center, and $E=1$, 1 CR radius to the right of the CR center (see bottom row of Figure \ref{fig:CR_images}).

\subsubsection{Evaluation on real eye-images}
How well does our approach perform on real eye images? To answer this question we tested our method against the thresholding and the radial symmetry methods when localizing the center of the CR in high resolution, high framerate videos of real eyes performing a collection of fixation tasks. Two different datasets were collected.

\paragraph{Dataset one}

\textit{Participants}. Eye videos were recorded from three participants. Two are authors of the current paper and the third is an experienced participant in fixation tasks. None of the participants wore glasses or contact lenses. Videos were recorded from the left eye. The study was approved by the Ethical Review Board in Sweden (Dnr: 2019-01081).

\textit{Apparatus}. The visual stimuli were presented on an ASUS VG248QE screen (531 x 299 mm; 1920 x 1080 pixels; \SI{60}{Hz} refresh rate) at a viewing distance of \SI{79}{cm}. 

Videos of the subject's left eye were acquired using our FLEX setup \cite{hooge2021pupil,Nystroem2022}, a self-built eye tracker. The setup consisted of a Basler camera (Basler Ace acA2500-60um) fitted with a \SI{50}{mm} lens (AZURE-5022ML12M) and a Near-IR Long pass Filter (MIDOPT LP715-37.5). Eye videos were recorded at \SI{500}{Hz} at a resolution of 896 × 600 pixels (exposure time: \SI{1876}{\micro\second}, Gain: 10 dB), and were streamed into mp4-files with custom software using libavcodec (ffmpeg) 5.1.2 and the libx264 h.264 encoder (preset: veryfast, crf: 0 (lossless), pixel format: gray). Videos were acquired and stored at 8-bit luminance resolution. The EyeLink \SI{890}{nm} illuminator was used (at 75\% power) to deliver illumination to the eye and to generate a reflection on the cornea that can be tracked in the eye image. An example eye image is shown in the left panel of Figure \ref{fig:input_eye_images}.

\textit{Procedure}. The subjects performed the following tasks where they looked at a fixation point consisting of a blue disk (\SI{1.2}{\degree} diameter), with a red point (\SI{0.2}{\degree} diameter) overlaid:

\begin{figure*}%
\centering
\subfigure{\includegraphics[height=0.25\textwidth]{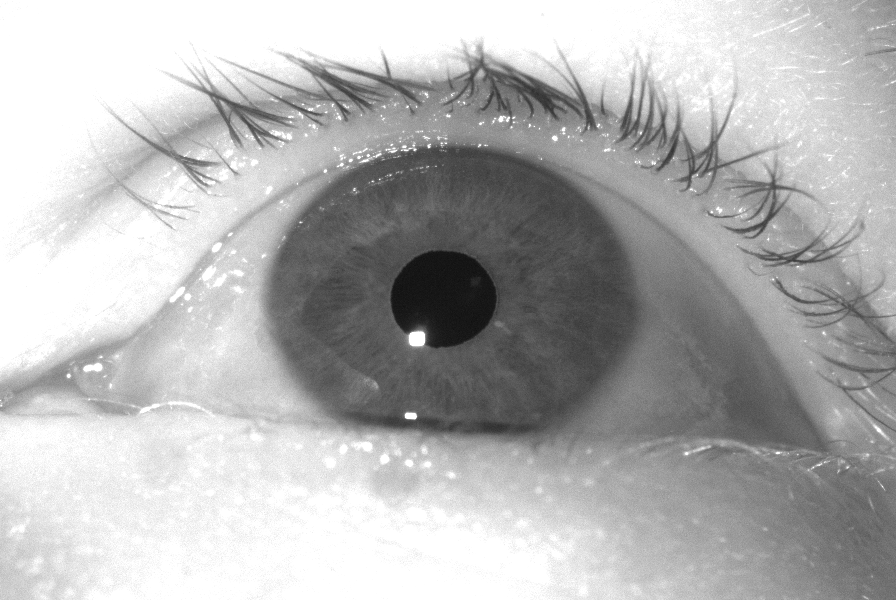}}
\subfigure{\includegraphics[height=0.25\textwidth]{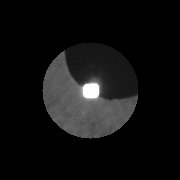}}
\caption{Full eye image (left) and masked cutout as processed by the radial symmetry and CNN methods (right).}%
\label{fig:input_eye_images}
\end{figure*}

\paragraph{Dataset two}

An abbreviated protocol was used to collect additional eye videos from the left eye of 17 participants (age 30-61 yrs (mean 45.4 yrs), five females, eleven males, one non-binary) who did not wear glasses or contact lenses. The study was approved by the Ethical Review Board in Sweden (Dnr: 2019-01081). Two are authors of the current paper. One participant was excluded because an impurity on their cornea caused an additional corneal reflection that none of the examined methods could handle.

To further examine the robustness of our method to variations in the luminance profile of the input eye images, this second recording was performed with the FLEX setup set to acquire images at \SI{1000}{Hz}. The captured eye images were less bright at this higher sampling rate due to the shorter possible exposure time. The videos were acquired at a resolution of 672 × 340 pixels (exposure time: \SI{882}{\micro\second}, Gain: 12 dB). The EyeLink \SI{890}{nm} illuminator was used (at 100\% power) to deliver illumination to the eye.

This data collection used the same displays as the previous, and the reduced protocol consisted of:
\begin{enumerate}
    \item Nine 1-second fixations on a 3×3 grid of fixation points positioned at $h = \{-7,\allowbreak 0,\allowbreak 7\}$ deg and $v = \{-5,\allowbreak 0,\allowbreak 5\}$ deg in random order.
    \item One 30-second fixation on a dot that was presented at $(0, 0)$ deg on a middle gray background.
    \item Two blocks of fifteen 1.5-second fixations on a 5×3 grid of fixation points positioned at $h = \{-7,\allowbreak -3.5,\allowbreak 0,\allowbreak 3.5,\allowbreak 7\}$ deg and $v = \{-5,\allowbreak 0,\allowbreak 5\}$ deg. Fixation locations were randomly ordered within each block.
\end{enumerate}

\paragraph{Image analysis}

Image analysis was performed frame-wise. A first stage was performed using the steps described in \citet{Nystroem2022}. Briefly, an analysis ROI and fixed pupil and CR thresholds were set manually for each participant's videos to identify the pupil and CR in the images, as is commonly performed \cite{perez2003precise,sanagustin2010ITUgazetracker,barsingerhorn2018stereoET,zimmermann2016oculomatic,ivanchenko2021lowcost,hosp2020remoteeye}. We ran the analyses at different CR and pupil thresholds and selected the thresholds that maximized the precision of the signals. These thresholds were used to binarize the images and after morphological operations to fill holes, the pupil and CR were selected based on shape and size criteria. The center of the pupil and CR were then computed as the center of mass of the binary blobs. The CR center provided by this method will be referred to as the CR center localized using the thresholding method.

In a second stage, a 180×180 pixel cutout centered on the center location identified by the thresholding method was made. A black circular mask with a radius of 48 pixels (about three times the horizontal size of the CR blob) was furthermore applied to the input image (see right panel in Figure \ref{fig:input_eye_images}). These masked images were then fed into the radial symmetry and CNN methods and their indicated CR centers stored.

To assess whether our method also works on lower resolution eye images as may be delivered by other eye tracking setups, we reran the image analysis described above with all input images downsampled by a factor of 2. The processing method and parameters were identical to those for the full resolution eye videos, except that the radius of the black circular mask applied to the CNN's input images was also halved.

\paragraph{Data analysis}
To investigate the data quality of the resulting signals, the following metrics were calculated.

First, RMS-S2S precision \cite{HoNyMu2012,niehorster2020characterizing,niehorster2020impact} of the CR center signals estimated using the three methods was computed in camera pixels for all the collected gaze data using a moving 200-ms window, after which for each trial the median RMS from all these windows was taken \cite{niehorster2020glassesviewer, hooge2018human, hooge2022robust}. The same calculation was performed for the pupil center signal.

Then, the accuracy, RMS-S2S precision of the calibrated gaze signal computed based on the three CR center signals were estimated. For this, gaze location was determined using standard P--CR methods: after subtracting the CR center location from the pupil center location, the resulting P--CR gaze data were calibrated using the gaze data collected on the 3×3 grid of the first task. Calibration was performed with second-order polynomials in $x$ and $y$ including first-order interactions \cite{stampe1993heuristic,cerrolaza2012error}:
\begin{equation}
p_{gaze}=a+bx+cy+dx^2+ey^2+fxy,
\end{equation}
where $p_{gaze}$ is the gaze position in degrees. The same formula was applied to compute the horizontal and vertical gaze positions. The accuracy of the gaze signal was then computed for each trial as the offset between the median estimated gaze location and the fixation point location for the data of task 4 in dataset one and task 3 in dataset two. The accuracy values for the repeated fixations on the 15 fixation targets were averaged. Similarly to the CR and pupil center signals, RMS-S2S and also STD precision of the gaze signal was computed in moving 200-ms windows for all the collected gaze data, after which for each trial the median RMS or STD value from all these windows was taken. 

\section{Results}

\subsection{Optimal CR center localization performance}
\begin{figure*}%
\centering
\subfigure{\includegraphics[width=0.3138\textwidth]{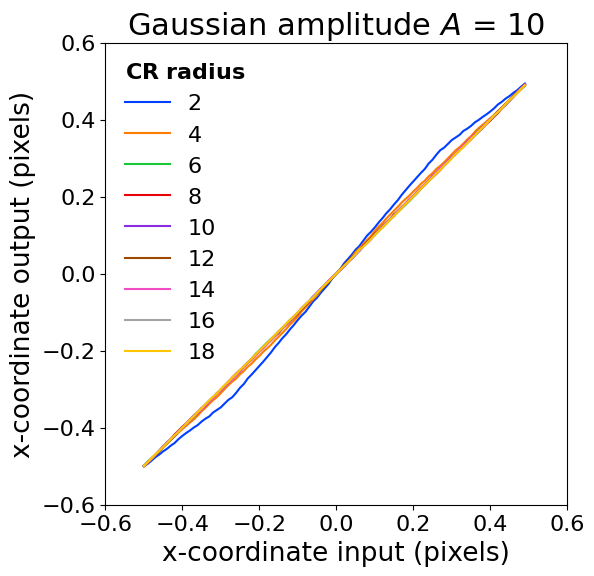}}
\subfigure{\includegraphics[width=0.2778\textwidth]{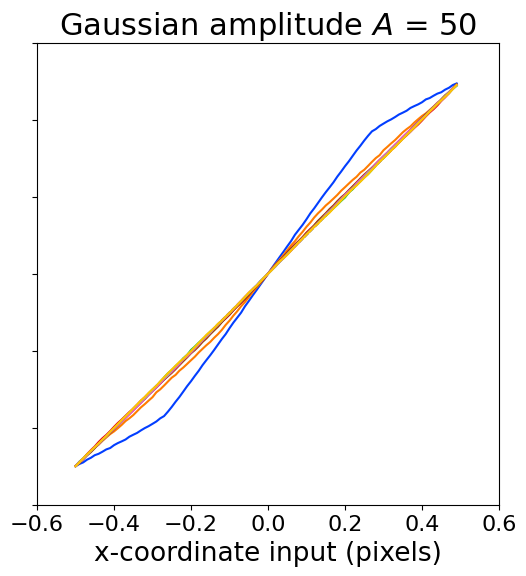}}\\
\subfigure{\includegraphics[width=0.3138\textwidth]{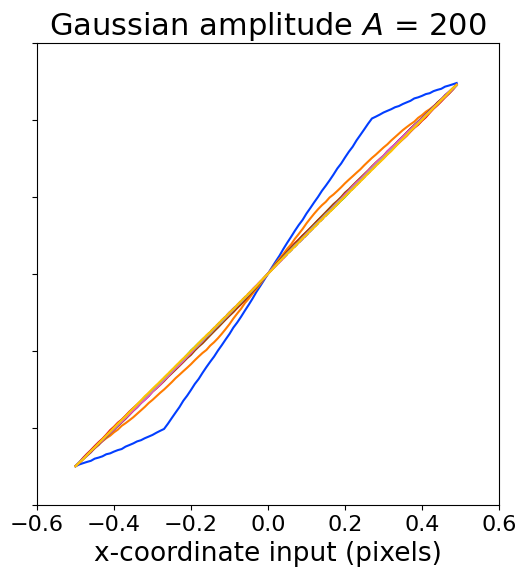}}
\subfigure{\includegraphics[width=0.2778\textwidth]{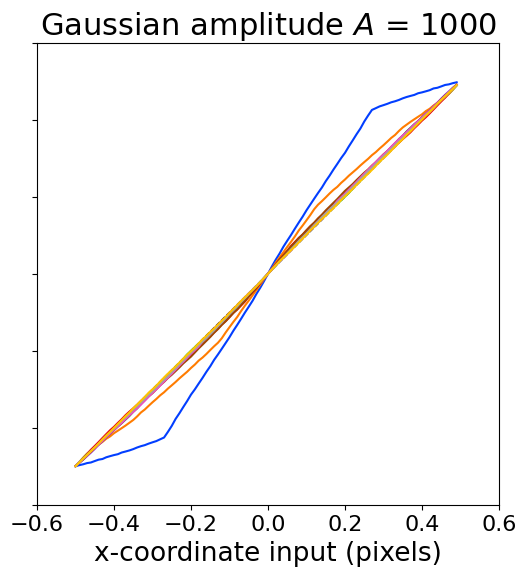}}
\subfigure{\includegraphics[width=0.2778\textwidth]{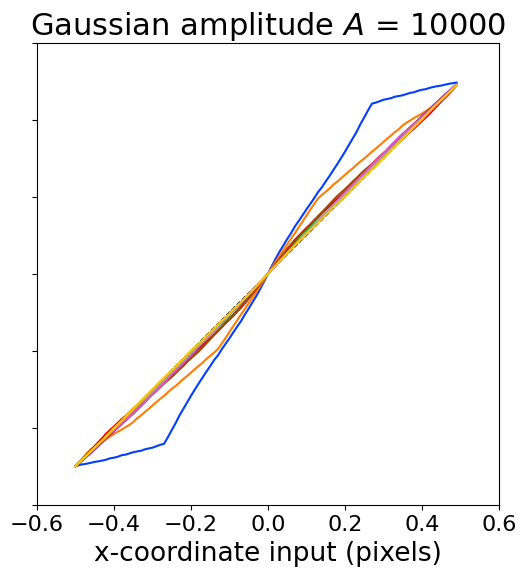}}
\caption{Best achievable CR center localization errors for different Gaussian amplitudes $A$ (different panels) and CR radii $r$ (different lines in each panel).}%
\label{fig:optimal_perf}
\end{figure*}

Figure \ref{fig:optimal_perf} shows the best obtainable CR center localization performance based on the information in the synthetic CR images for the different Gaussian amplitudes (different panels) and CR sizes (lines within each panel). As can be seen, appreciable errors in CR center location occur at all examined Gaussian amplitudes for the smallest CR size (2), and also for CR size 4 for higher Gaussian amplitudes (i.e. images containing narrower tails). Furthermore, error increases as a function of Gaussian amplitude (narrower tails). To illustrate how close the different CR center localization methods are to their optimal performance, the results of this examination will be used as reference lines when presenting the evaluation on synthetic images in the ``\nameref{sec:eval_sim}'' section below.

\subsection{Evaluation on synthetic CRs}
\label{sec:eval_sim}

\begin{figure*}%
\centering
\subfigure{\includegraphics[height=0.35\textwidth]{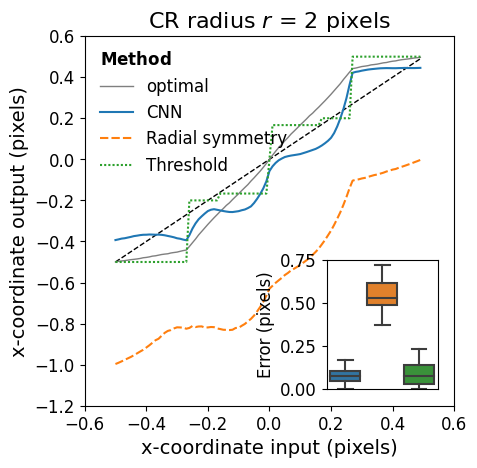}}
\subfigure{\includegraphics[height=0.35\textwidth]{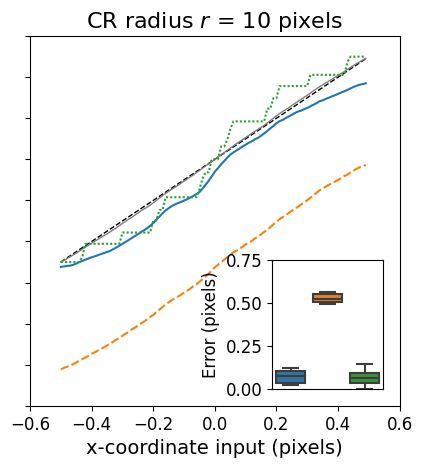}}
\subfigure{\includegraphics[height=0.35\textwidth]{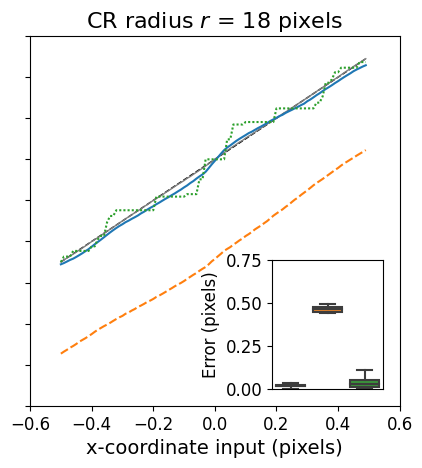}}
\caption{Errors in CR center localization for different CR sizes $r$ for three methods. The panel insets show boxplots of
the CR center localization error for each estimated input position. For all these simulations, $A=10000$, $E=0$, $I=128$}%
\label{fig:results_error_traces}
\end{figure*}

First we set out to examine whether our method was able to locate the CR center more accurately than two commonly used algorithmic approaches when applied to synthetic data.
Figure \ref{fig:results_error_traces} shows the error in CR center localization achieved by the three methods for three different CR sizes. Negative errors are leftward, and positive rightward. For illustration purposes, results are shown for Gaussian amplitude $A=10000$ and a half-grey background ($E=0$, $I=128$). As can be seen, for the smallest CR size, the CNN and thresholding methods perform similarly, while the radial symmetry method shows a larger bias towards the left of the image, which is the gray side. As CR size increases, this bias towards the grey side of the image for the radial symmetry method only slightly decreases. For these larger CR sizes, the center localization output of the threshold and CNN methods becomes more smooth, and the CNN by and large shows a lower error than the threshold method.

\begin{figure*}%
\centering
\subfigure{\includegraphics[height=0.3818\textwidth]{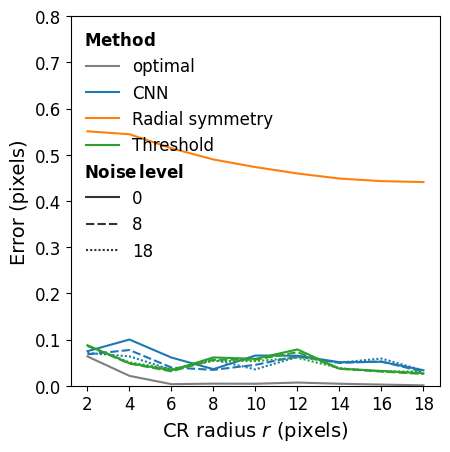}}
\subfigure{\includegraphics[height=0.375\textwidth]{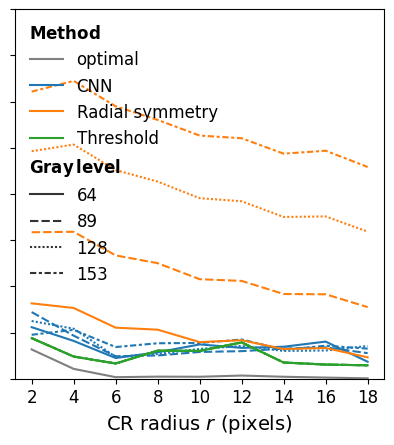}}\\
\subfigure{\includegraphics[height=0.3818\textwidth]{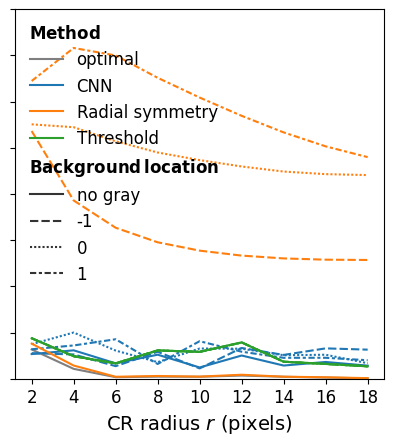}}
\subfigure{\includegraphics[height=0.375\textwidth]{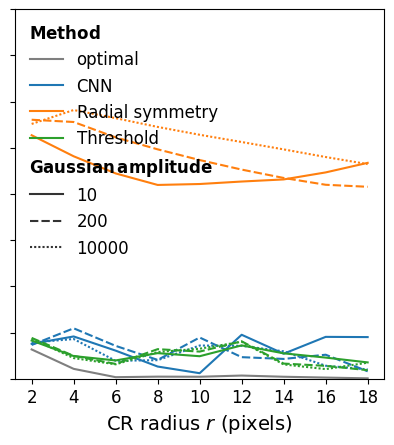}}
\caption{Errors in CR center localization for the three methods as a function of CR radius, for different noise levels (top-left), pixel intensity levels of the lighter background section (top-right), locations of the gray background (bottom-left) and Gaussian amplitudes (bottom-right). For the top-left panel, average error of the radial symmetry method ranged from 3--8 pixels at noise level 18, not shown. For the top-right and bottom panels, the noise level was 0. For the bottom panels, the pixel intensity level of the lighter background section $I$ was 128. For the bottom-right panel, the background location $E$ was 0.}%
\label{fig:results_error_avg}
\end{figure*}

Localization performance of the three methods as a function of CR size for three different pixel noise levels is plotted in Figure \ref{fig:results_error_avg} (top-left panel). As can be seen, error in localization is almost independent of CR size for the threshold and CNN methods while error decreases as a function of CR size for the radial symmetry method. The thresholding and CNN methods are not affected by noise in the input image in the range that we examined and achieve similar CR center localization error. On the other hand, the radial symmetry method was strongly affected by pixel noise level (errors were mostly over 0.5 pixels at noise level 8 and ranged from 3--8 pixels at noise level 18, not shown). As such, further plots show results at noise level 0 to highlight the best possible performance of the radial symmetry method.

The effect of the pixel intensity level of the lighter background section is shown in Figure \ref{fig:results_error_avg} (top-right panel). Similar to the effect of pixel noise level, the localization error of the threshold and CNN methods, but not the radial symmetry method, is almost independent of CR size and background pixel intensity level. Furthermore, performance of the CNN method is very close to that of the threshold method, with both achieving errors of around to well below 0.1 pixels across CR sizes.

The effect of the location of the grey background on localization performance is shown in Figure \ref{fig:results_error_avg} (bottom-left panel). While all methods performed nearly perfectly when the background was fully black, only the threshold and CNN methods are stable over different locations of the gray background. The performance of the radial symmetry method is strongly affected by the position of the gray background, and overall shows much larger errors than when no gray background was present.

The effect of the width of the tail of the CR is shown in Figure \ref{fig:results_error_avg} (bottom-right panel). Recall that more saturated Gaussians (those with larger amplitude $A$) have narrower tails (c.f. Figure \ref{fig:CR_images}). As can be seen, the effect of tail width on CR localization performance is minimal for the three methods. Taken together, importantly, the localization performance of the CNN is on par with the best-performing algorithmic approach to CR localization, achieving average errors of around or well below 0.1 pixels in all cases.

\subsection{Evaluation on real eye images.}

\begin{figure*}%
\centering
\subfigure{\includegraphics[height=0.35\textwidth]{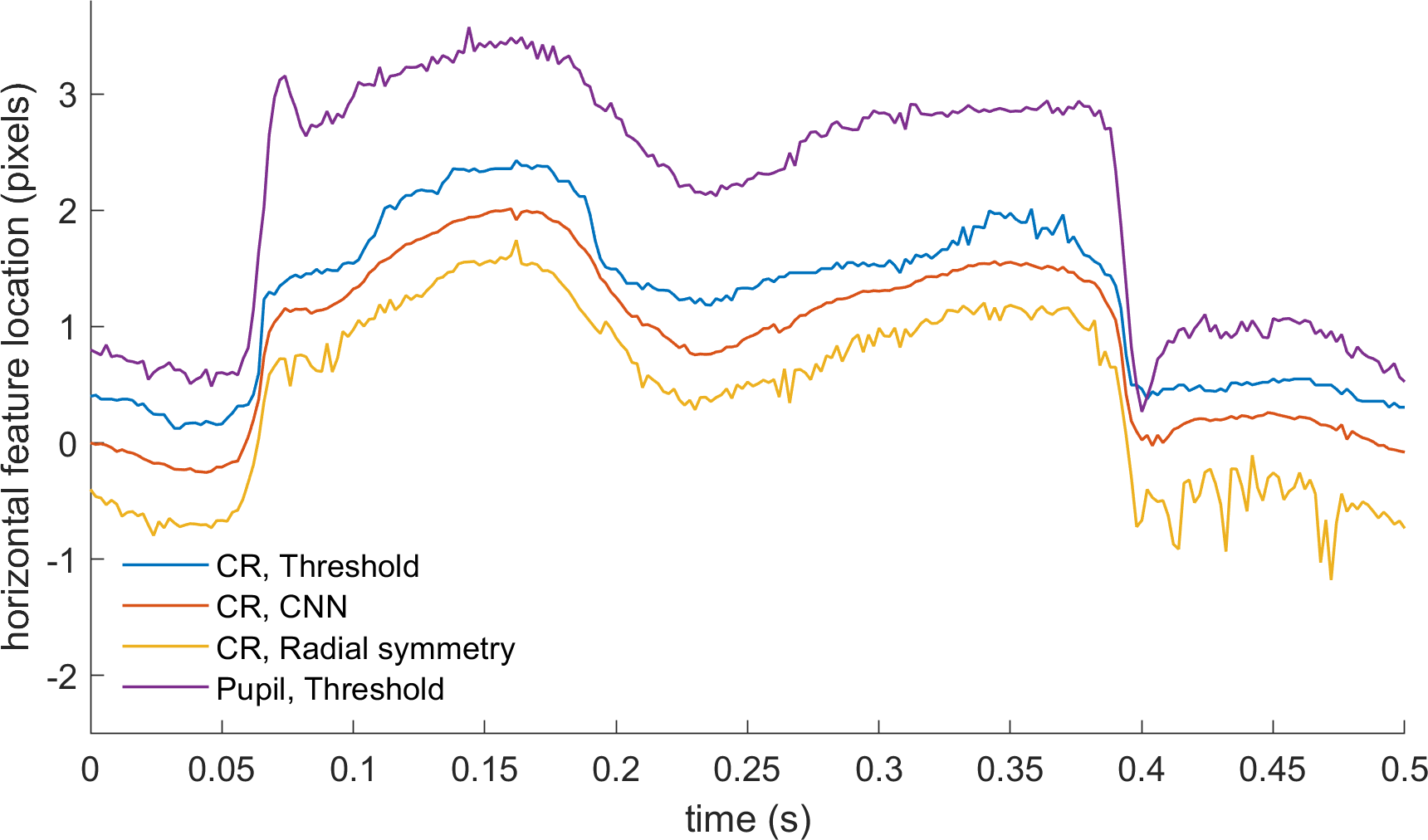}}
\subfigure{\includegraphics[height=0.35\textwidth]{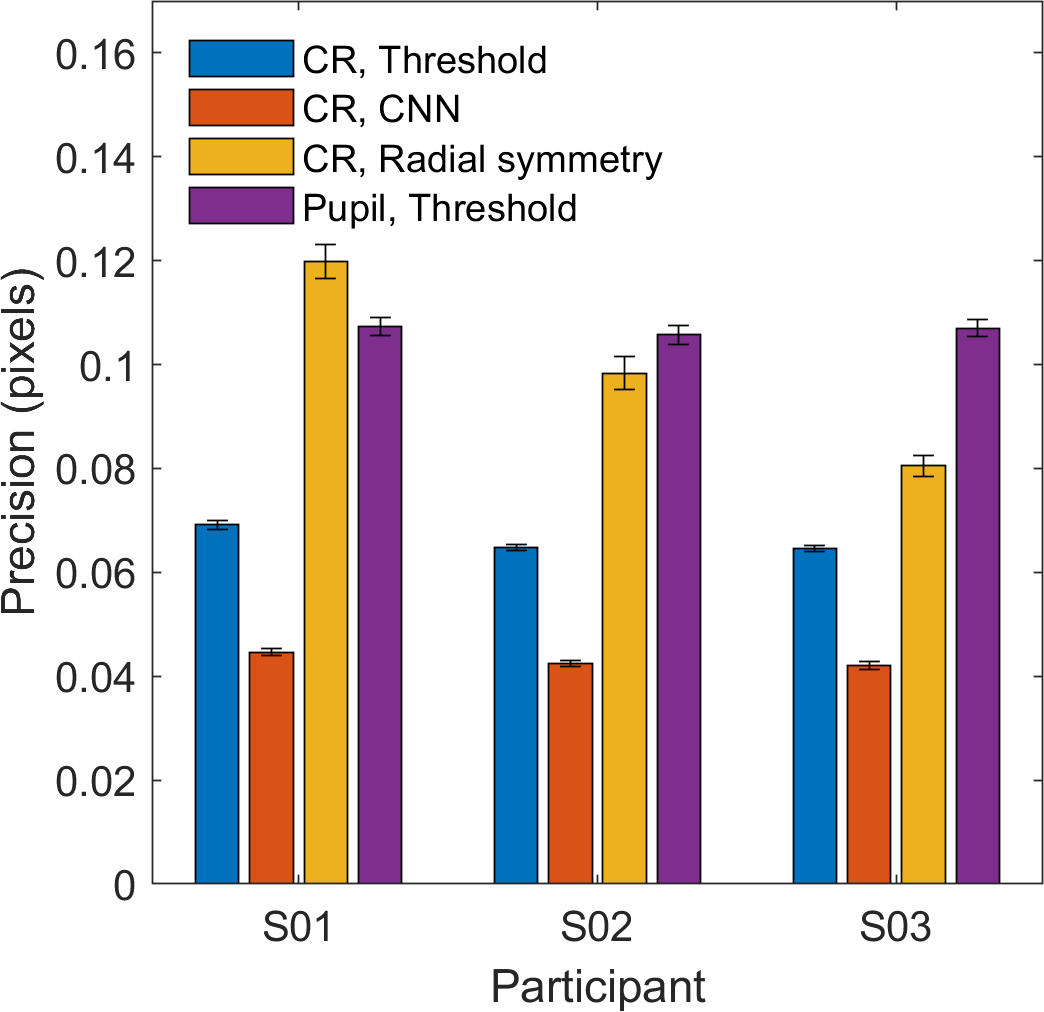}}
\caption{Real eye CR and pupil center signals of dataset one}. Left: representative segment of pupil and CR center signals from S03 in camera pixels. For the CR center, the signals produced by three different CR center localization methods are shown. The signals contain two small saccades and have been vertically offset for clarity. RMS precision for the shown segments are \SI{0.081}{\pixel} for the Threshold signal, \SI{0.061}{\pixel} for CNN, and \SI{0.096}{\pixel} for Radial symmetry and \SI{0.121}{\pixel} for the pupil signal. Further, an RMS precision comparison (right panel) between the three methods and the pupil signal on all data of three participants is shown. Error bars depict standard error of the mean.%

\label{fig:real_eye_results_raw}
\end{figure*}

\subsubsection{Dataset one}
Next, we set out to test whether our method is able to perform CR center localization in real eye images and if so, whether it delivers a CR position signal with higher precision than two algorithmic approaches.
To test how well our method works on real eye images, we first performed CR and pupil center localization on dataset one, which consisted of \SI{500}{Hz} videos of eye movements made by three participants. CR localization was performed by three methods.

The left panel of Figure \ref{fig:real_eye_results_raw} shows an example segment of CR center locations estimated using the three methods, along with the estimated pupil center location. A can be seen, the CR center signal from the CNN method appears smoother than the signal from the threshold method, while the signal from the radial symmetry method looks less smooth than the threshold signal. The pupil center signal by and large looks similarly noisy as the CR signal from the radial symmetry method.

To quantify these observations, we calculated the RMS precision of all four signals for all recorded videos of three participants. The results of this analysis are shown in Figure~\ref{fig:real_eye_results_raw} (right panel). While there were differences in overall noise level between participants, a clear pattern in results for the CR center localization methods is seen. The CNN method consistently delivers signals with a better precision (lower values) than the thresholding method, while the radial symmetry method delivers signals with worse precision (higher values). Precision of the pupil center signal is consistently much worse than that of the CNN- or thresholding-based CR center signals. It is important to note here that all methods processed each video frame independently, and that improved precision could thus not be due to any form of temporal information being used from previous or future frames \cite[c.f.][]{niehorster2020apparent,niehorster2020characterizing}.

\begin{figure*}%
\centering
\subfigure{\includegraphics[height=0.35\textwidth]{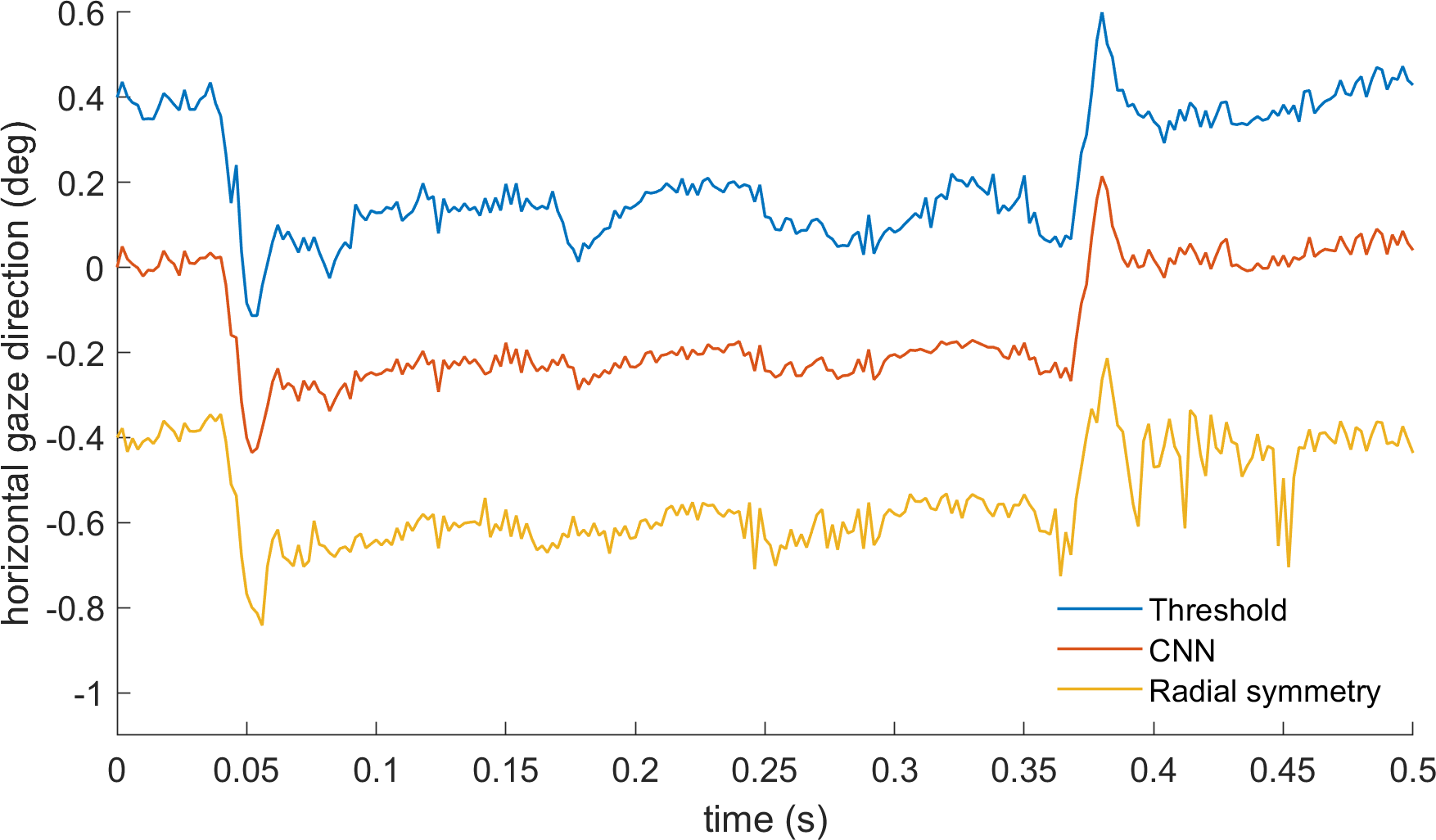}}\\
\subfigure{\includegraphics[width=0.31\textwidth]{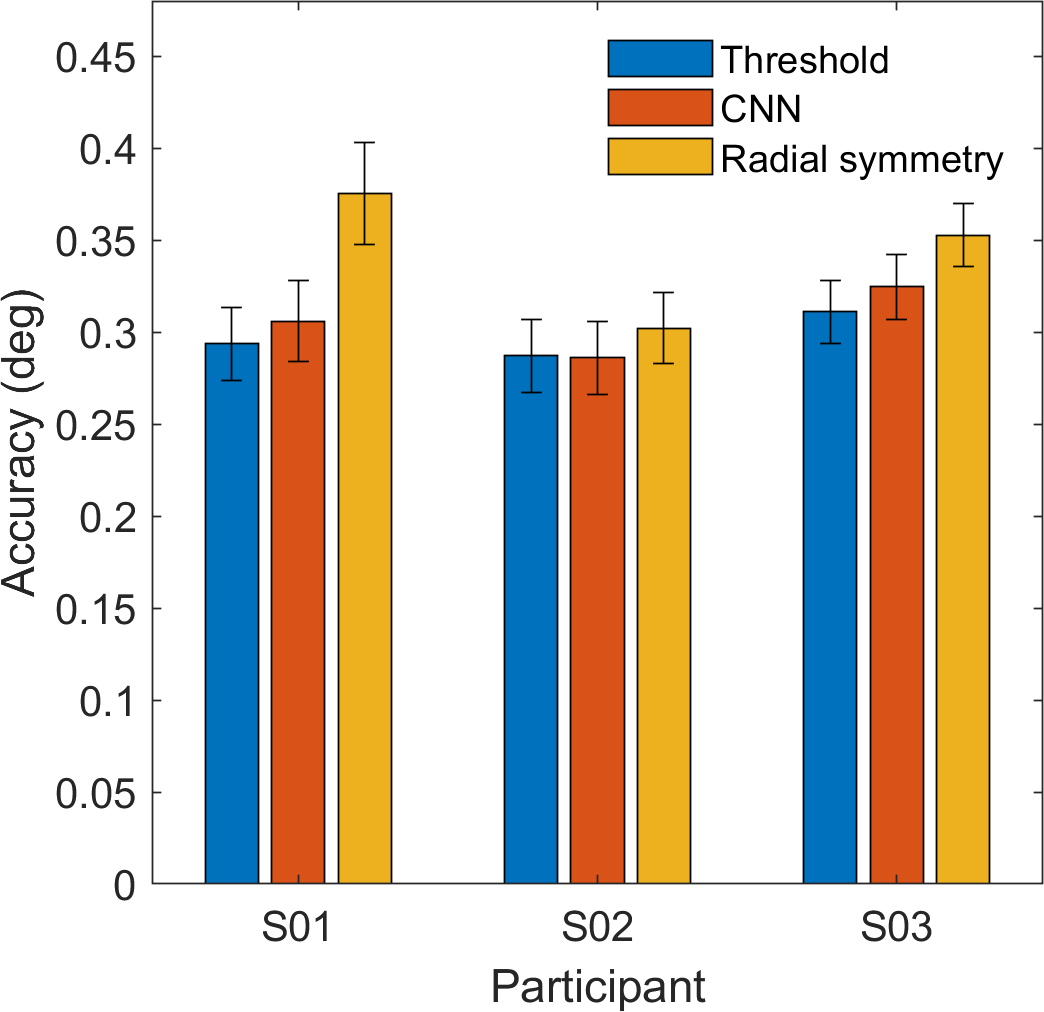}}
\subfigure{\includegraphics[width=0.31\textwidth]{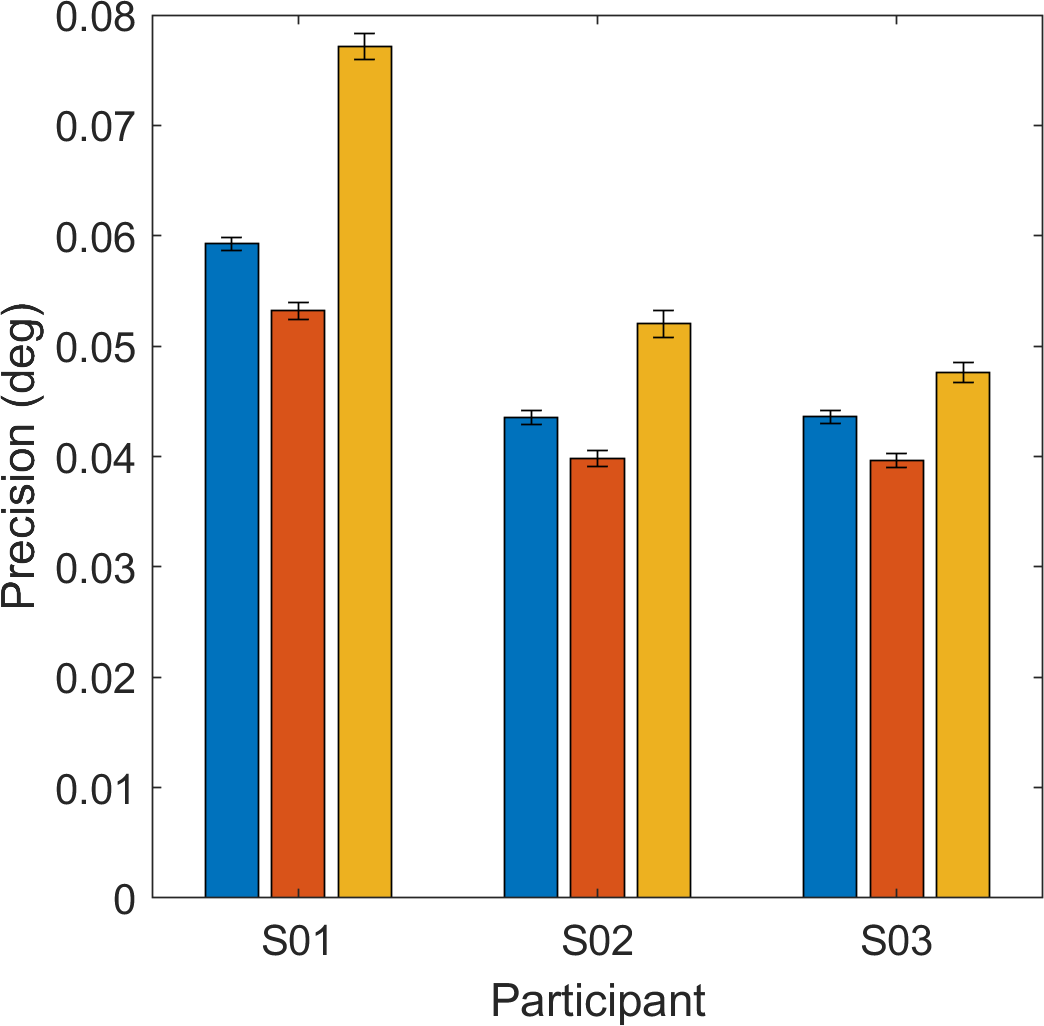}}
\subfigure{\includegraphics[width=0.31\textwidth]{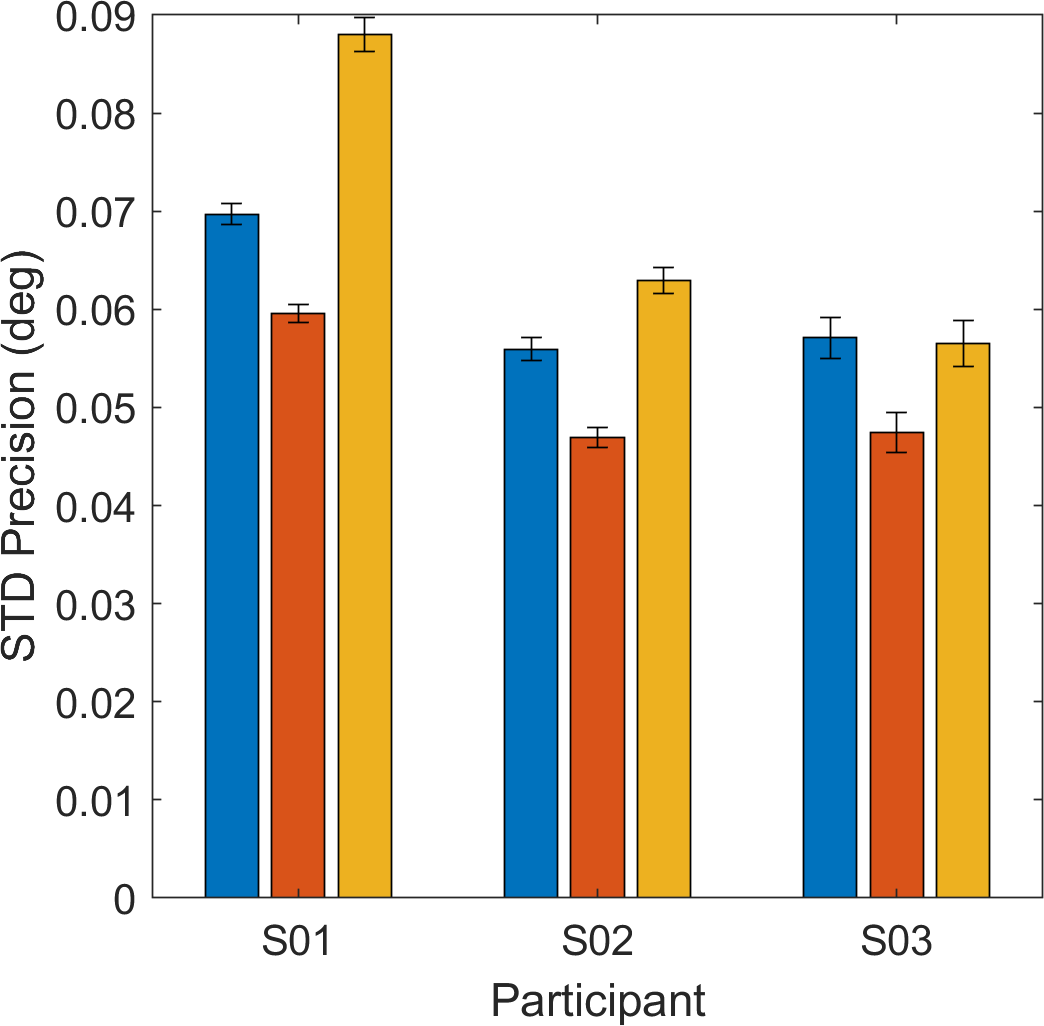}}
\caption{Real eye calibrated gaze signals of dataset one}. Top: representative segment of calibrated P-CR signals from S03 as processed by three different CR center localization methods. The signals contain two small saccades and have been vertically offset for clarity. RMS precision for the shown segments are \SI{0.040}{\degree} for the Threshold signal, \SI{0.034}{\degree} for CNN, and \SI{0.046}{\degree} for Radial symmetry. Further, an accuracy comparison (bottom left panel), an RMS precision comparison (bottom middle panel) and an STD precision comparison (bottom right panel) between the three methods on data of three participants are shown. Error bars depict standard error of the mean.

%
\label{fig:real_eye_results_cal}
\end{figure*}

How does the improved CR center localization of our method impact the gaze signal? To answer this question, we performed a similar analysis as above, but on the calibrated gaze signals. The top panel of Figure \ref{fig:real_eye_results_cal} shows an example segment of gaze data computed from the three signals. As can be seen, the gaze signals derived from the three different CR localization methods look much more similar than the CR center signals in Figure \ref{fig:real_eye_results_raw} (left panel). This is likely due to the fact that derivation of the gaze signal involves subtracting the estimated CR center location from the much noisier pupil center location estimate \cite[c.f.][]{niehorster2018rawclassify}. The noise in the pupil center location estimate likely is the dominant component of noise in the derived gaze signal, to a large extent swamping the differences in precision between the CR center location signals.

The bottom panels of Figure \ref{fig:real_eye_results_cal} show the accuracy, RMS-S2S, and STD precision achieved with the three CR center localization methods. While there were small differences between the participants, no systematic differences in accuracy between the three methods were observed. Overall, both the RMS-S2S and the STD precision of the gaze signals derived from the CR center localization estimates of the CNN was a little lower than for the gaze signal derived from the threshold-based CR center, while that for the gaze signal derived from the radial symmetry method for CR center localization showed worse precision.

\subsubsection{Dataset two}

\begin{figure*}%
\centering
\subfigure{\includegraphics[height=0.35\textwidth]{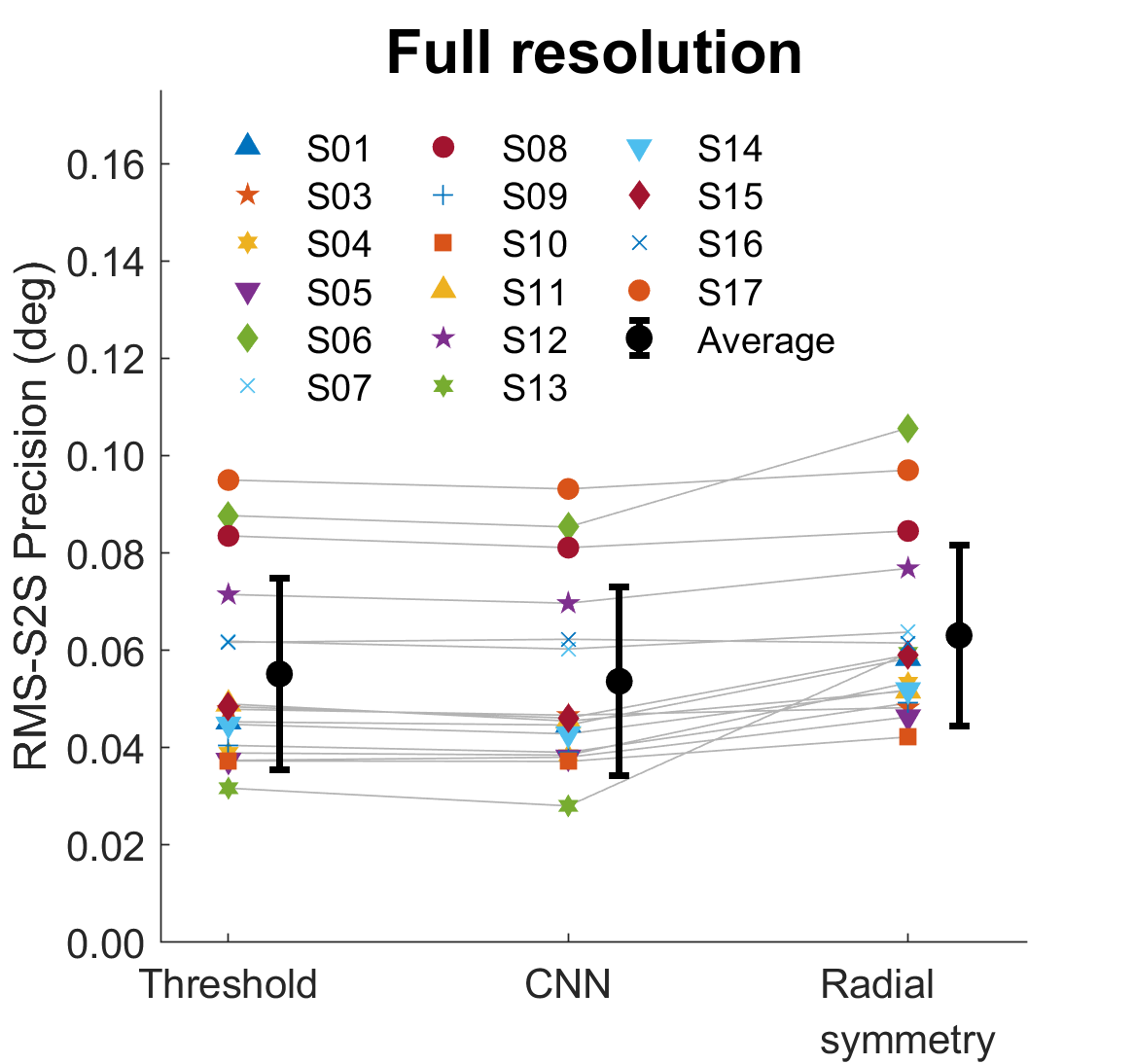}}
\subfigure{\includegraphics[height=0.35\textwidth]{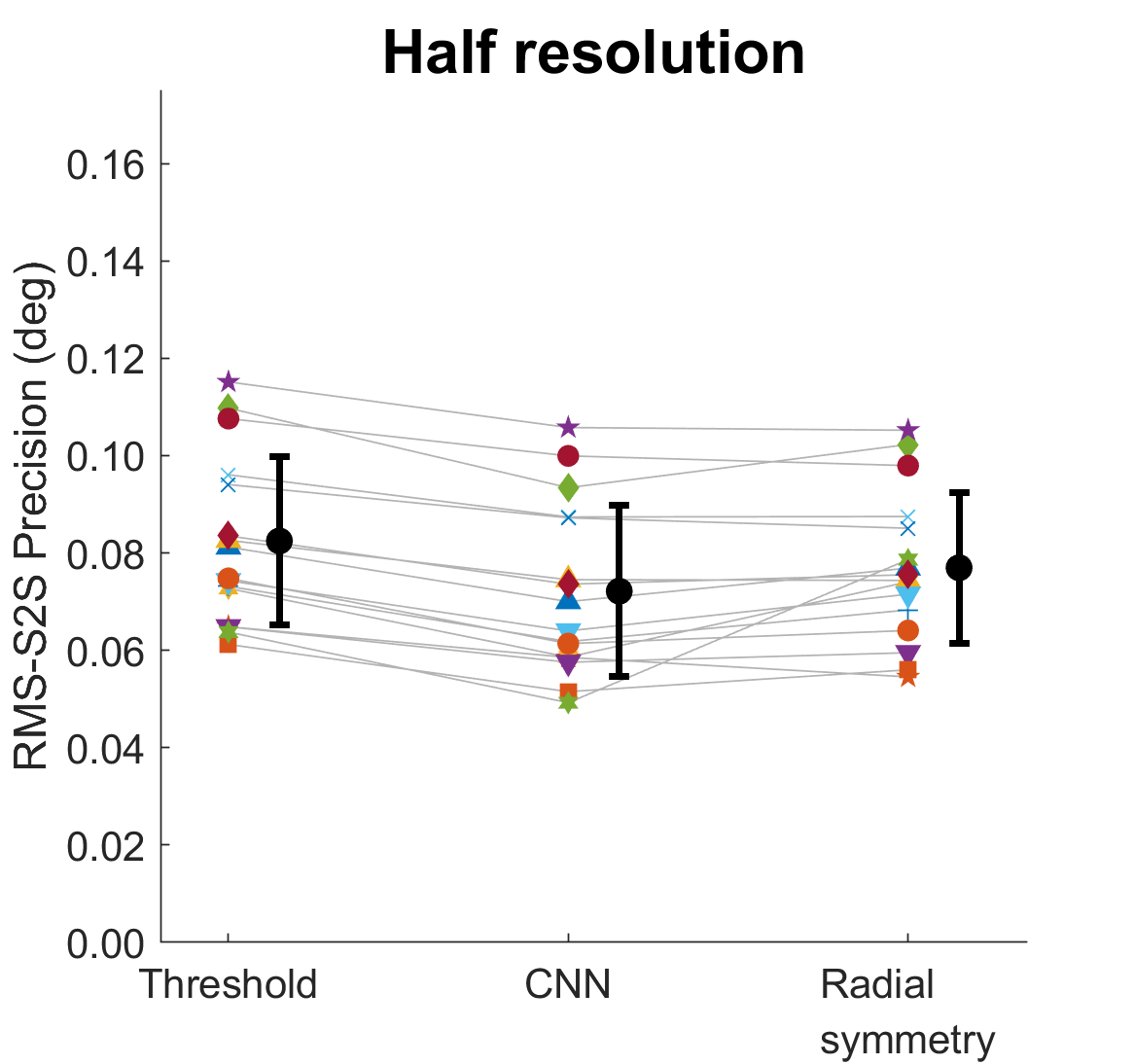}}
\caption{RMS-S2S precision of the raw signals for dataset two. An RMS precision comparison between the CR center signals derived from the three methods and the pupil center signal is shown for all participants (colored symbols) along with the mean across participants (black circles) for analyses run both at full video resolution (left panel) and at half resolution (right panel). Error bars depict standard error of the mean.}
\label{fig:real_eye_raw_d2}
\end{figure*}

To examine how our method performs on real eye images across a wider range of participants with different eye physiology and for lower resolution eye images, we have collected a new set of 17 participants (one of which was excluded from analysis, see methods) and analyzed the videos captured both at full and at half resolution.

Figure \ref{fig:real_eye_raw_d2} shows the calculated RMS precision of the CR signal processed using three different methods, and the pupil signal, for all recorded videos of all participants. The same pattern emerges for the full resolution and the half resolution videos. As we saw for dataset one, while there were differences in overall noise level between participants, a clear pattern of results emerges where the CNN method delivers signals with a better precision (lower values) than the thresholding method, and the radial symmetry method performs worse (higher values) than the CNN method. As before, the precision of the pupil center signal is worse than that of the CNN- or thresholding-based CR center signals. 

\begin{figure*}%
\centering
\subfigure{\includegraphics[height=0.35\textwidth]{paper_rms_d2.png}}
\subfigure{\includegraphics[height=0.35\textwidth]{paper_rms_d2_scale2.png}}\\
\subfigure{\includegraphics[height=0.35\textwidth]{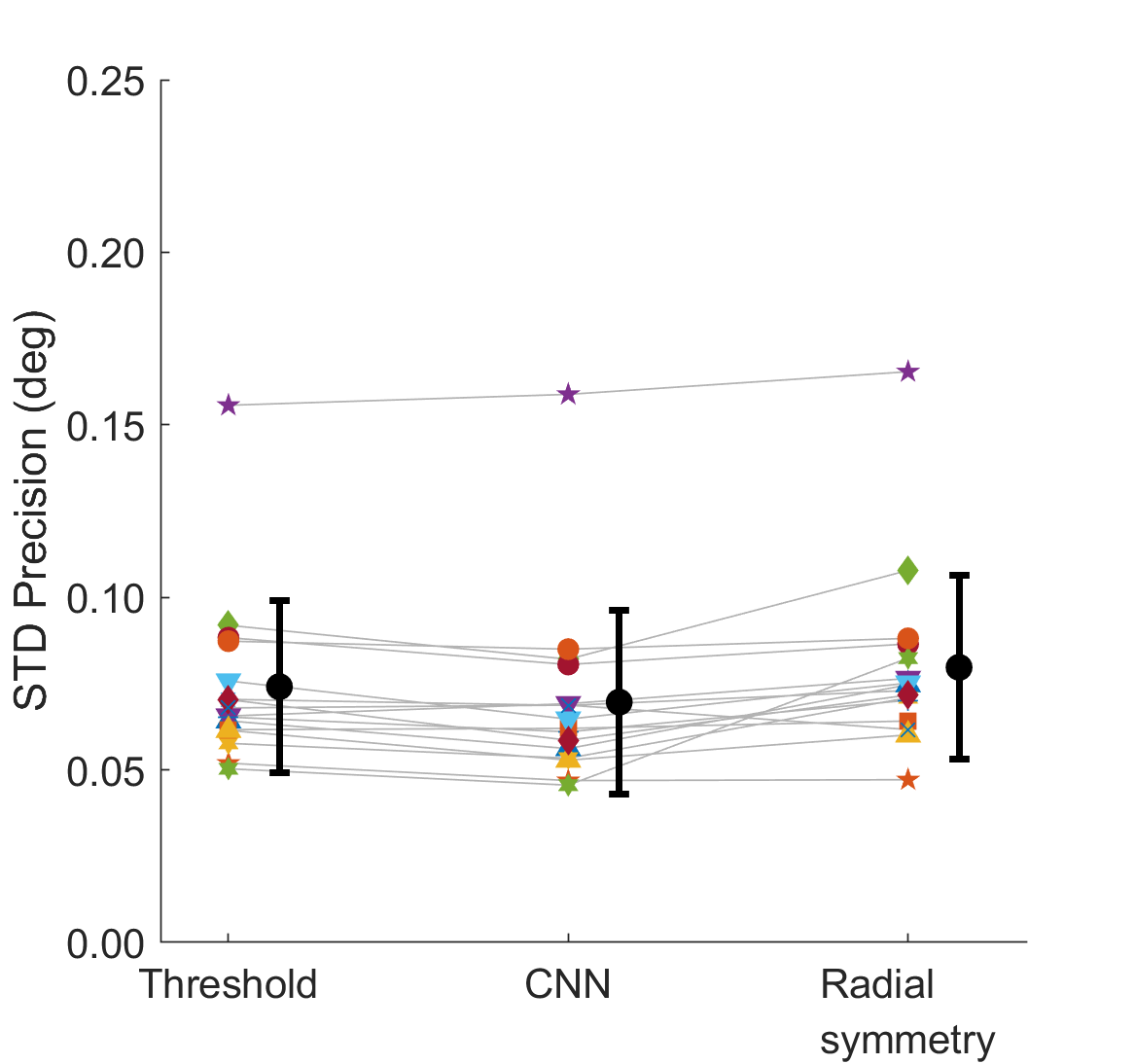}}
\subfigure{\includegraphics[height=0.35\textwidth]{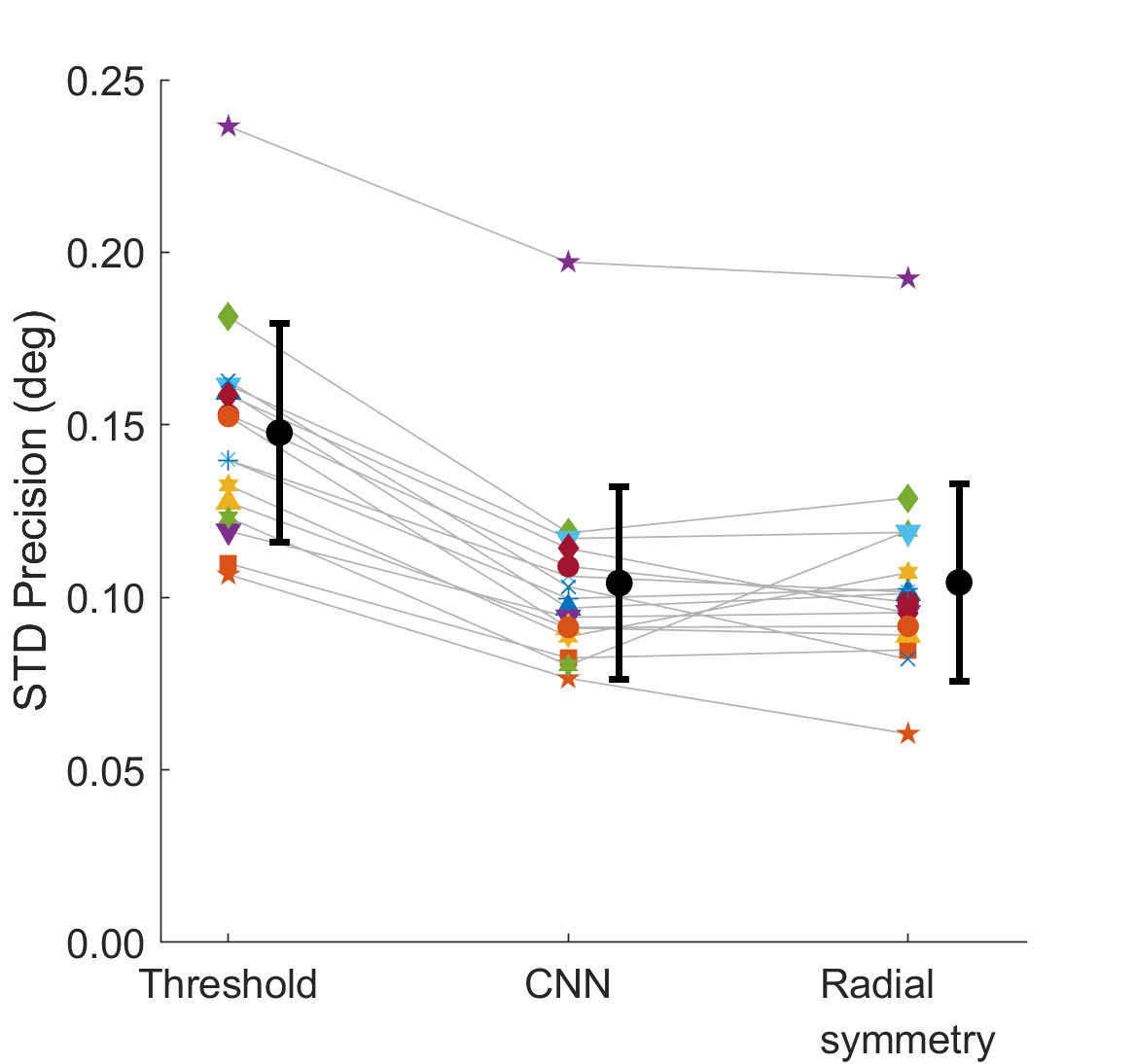}}\\
\subfigure{\includegraphics[height=0.35\textwidth]{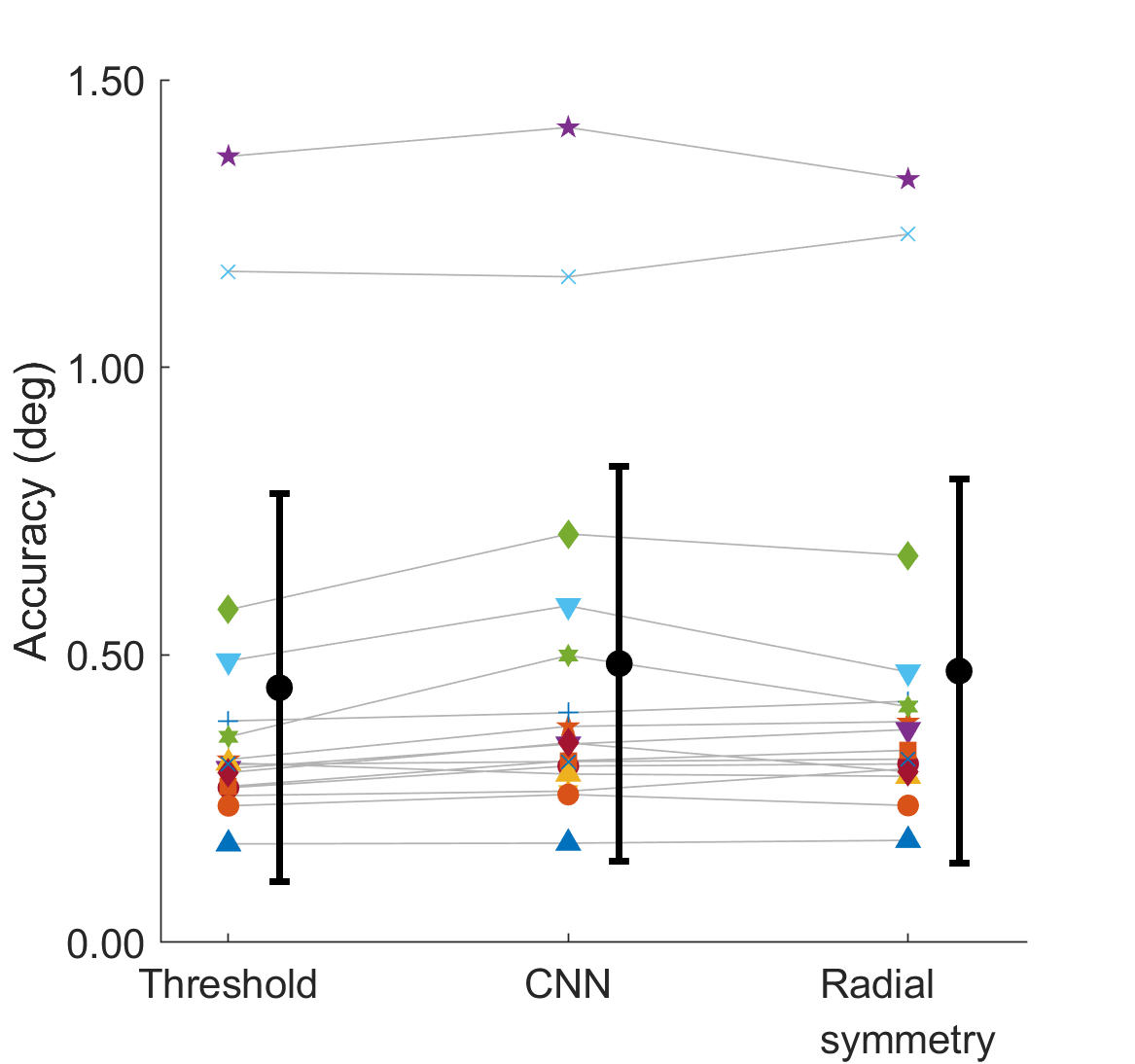}}
\subfigure{\includegraphics[height=0.35\textwidth]{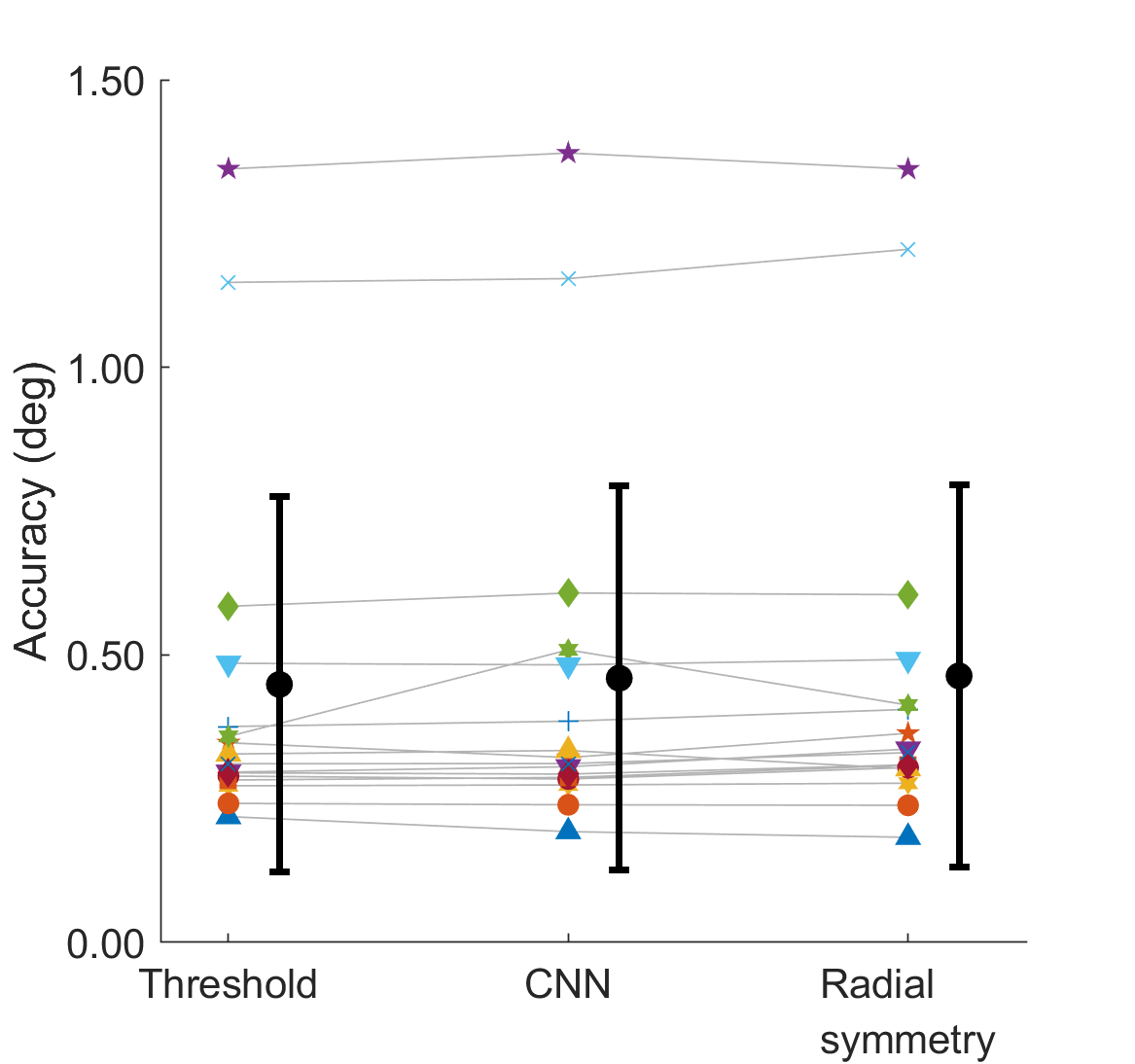}}
\caption{Data quality of calibrated gaze signals of dataset two. RMS-S2S precision (top panels), STD precision (middle panels) and accuracy (bottom panels) comparisons of the calibrated gaze signals derived from the three CR center localization methods is shown for all participants (colored symbols) along with the mean across participants (black circles) for analyses run both at full video resolution (left panels) and at half resolution (right panels). Error bars depict standard error of the mean.}
\label{fig:real_eye_d2}
\end{figure*}

To examine how the CR center localization methods impact the resulting calibrated gaze signals, we computed the RMS-S2S and STD precision, and the accuracy of the calibrated gaze signals of each participant for both the full resolution and half resolution video analyses. As for dataset one, in most cases there were only small differences in RMS-S2S and STD precision (Figure \ref{fig:real_eye_d2}, top and middle rows) between the three CR center localization methods for both video resolutions, with the CNN method showing slightly better precision (lower values) than the other methods. Only for the half-resolution analysis was the STD precision of the gaze signal derived from the threshold method clearly worse (higher values) than for the signals derived from the CNN and radial symmetry methods. Accuracy did not vary systematically between the three methods.

\section{Discussion}
In this paper, we have developed a CNN architecture and training regime for localizing single CRs in eye images. We have furthermore analyzed the spatial accuracy and precision obtained with this new method using synthetic and real eye images. Regarding the research questions posed in the introduction, the contributions of this paper are: that 1) we provide a simple method using only synthetic images to train a CNN to perform CR center localization and demonstrate that a CNN trained using this method can perform CR center localization in real eye images; 2) we show that our method is able to locate the CR center similarly accurately as a commonly used algorithmic approach when applied to synthetic data; and 3) we demonstrate that our method can outperform algorithmic approaches to CR center localization in terms of spatial precision when applied to real eye images.

Specifically, the paper has shown that our CNN-based method consistently outperforms the popular thresholding method for CR center localization as well as the radial symmetry method that was recently adopted by \citet{Wu2022}. As \citet{Nystroem2022} have recently shown, binarizing an eye image using a thresholding operation reduces the CR center localization accuracy compared to methods that use the full range of pixel intensity values in the image of the CR \cite[c.f. also][who show this in the context of microscopy]{helgadottir2019digital}. The radial symmetry method \cite{parthasarathy2012rapid} uses the full range of intensity values and has been shown to outperform thresholding for localization of the center of image features \cite{Wu2022,helgadottir2019digital,midtvedt2022single}. However, these results were obtained with features on a uniform background. Our simulations show that the radial symmetry method is consistently considerably worse when used on images with a background consisting of two regions with different luminance. 
It is therefore not suitable for use in more general eye tracking scenarios, where the CR is often overlaid on a non-uniform background, such as the iris or the edge of the pupil.
In contrast, our CNN was trained on highly simplified images that contained such backgrounds, and shows performance that is robust to their presence in both synthetic and real eye images. This demonstrates that the CNN approach, if appropriately trained, is able to use the pixel intensity information contained in the image of the CR to localize its center while effectively ignoring the background. Our CNN method consistently outperformed the other methods across evaluations performed on two different datasets and also when fed eye images that were downsampled to half resolution, showing that the method is applicable to many participants with differing eye physiology and generalizes to lower resolution eye images.

This paper has demonstrated that simple simulations can be used to effectively train deep learning models that work on real eye images, raising questions about the need for heavy data augmentation techniques and time-consuming data collection as well as hand labeling efforts or reconstruction methods. However, it is worth emphasizing that we have so far employed this approach only on high quality eye images (see Figure~\ref{fig:input_eye_images}) encountered in high-end lab-based eye tracking scenarios where researchers are interested in microsaccades and other fixational eye movements, as well as other aspects of eye movements that require high data quality, such as slow pursuit. While our approach shows promise for these research scenarios, other scenarios in which eye trackers are frequently applied such as virtual reality or wearable eye tracker settings face eye images of significantly worse quality. Our approach should thus be tested on more challenging targets (e.g. localizing the center of the pupil or iris), more complex situations (e.g. involving multiple CRs and spurious reflections) and images of lower quality to further test the hypothesis that effective gaze estimation methods in a broader context can be trained using simple simulations alone.

Our results show that while our method offered significantly reduced RMS precision in the CR center signal (\SIrange{28.0}{34.9}{\percent} lower than thresholding for dataset one, and on average \SI{13.0}{\percent} and \SI{41.5}{\percent} lower for the full and half resolution analyses of dataset two, respectively), this translated to an improvement in RMS precision of the gaze signal that ranged only between \SIrange{7.2}{8.6}{\percent} for dataset one, and on average \SI{2.9}{\percent} for the full resolution and \SI{13.0}{\percent} for the half resolution analysis of dataset two. Indeed, for a gaze signal that is derived using the P-CR principle, CR center localization performance is only half the story. P-CR eye trackers typically use the vector between the pupil and CR centers and as such noise in the pupil signal also plays an important role in determining the precision of the gaze signal. As shown in our results, for dataset one the noise in the pupil signal was between \SIrange{55}{66}{\percent} higher than in the CR center signal based on thresholding (average \SI{127}{\percent} at full and \SI{32.1}{\percent} at half resolution for dataset two). This ratio only worsens to between \SIrange{126}{151}{\percent} when considering the CR center signal produced by our CNN method for dataset one (average \SI{165}{\percent} at full and \SI{130}{\percent} at half resolution for dataset two). As such, further improvements in CR center localization precision will be of little practical use for P-CR eye trackers until the precision of pupil center localization is also improved.

In summary, our results indicate that our method for training deep learning models for eye tracking applications using only simple synthetic images shows great promise. However, much of the road ahead to a fully deep-learning based eye tracking method trained using only synthetic images remains unexplored. As the next endeavor, we plan to extend our approach to localization of the pupil center. If successful, this will not only provide a much needed improvement in precision of the pupil signal that will translate into increased quality of the P-CR gaze signal, but will also provide a further and more ambitious test of the hypothesis that effective gaze estimation methods can be trained using simple simulations alone. Why does localizing the pupil center provide a more ambitious test? While a CR has approximately the same size, shape and pixel intensity profile in all eye images for a given eye tracking setup and is thus easy to design a representative simulation for, this is not the case for the pupil. The luminance of the pupil in the eye image can vary significantly, and its apparent shape can change radically as it changes size, is imaged from different angles when the eye rotates and one or multiple CRs overlap it.

We further plan to develop our framework to handle more challenging tasks such as detecting multiple CRs and matching their positions in the eye image to the corresponding physical configuration of light sources, similar to the work in~\citet{chugh2021detection} and \citet{niu2021real}. This would enable using our method to be used in eye tracking scenarios outside of laboratory settings. 

\section{Acknowledgments}
We thank Ignace Hooge for helpful discussion. We gratefully acknowledge the Lund University Humanities Lab where data were recorded.

\section{Open practices statement}
The experiments were not preregistered. We have made our simulation and model training code, the trained model and the code for evaluating the model on simulated and real images available at the following link: \url{https://github.com/dcnieho/Byrneetal_CR_CNN}.

\printbibliography
 
\end{document}